\pdfoutput=1

\documentclass[11pt]{article}

\usepackage[final]{acl}

\usepackage{times}
\usepackage{latexsym}
\usepackage{subcaption}
\usepackage{amsmath}
\usepackage{booktabs}
\usepackage{float}
\usepackage[T1]{fontenc}

\usepackage[utf8]{inputenc}

\usepackage{microtype}
\usepackage{url}

\usepackage[autostyle=true]{csquotes}

\usepackage{inconsolata}

\usepackage{graphicx}

\usepackage{todonotes}

\usepackage{amsfonts}
\usepackage{amssymb}

\title{DS-ProGen: A Dual-Structure Deep Language Model for \\ Functional Protein Design}

\author{
\bf Yanting Li$^{\heartsuit}$$^{\spadesuit}$$^{\ast}$,
Jiyue Jiang$^{\heartsuit}$\thanks{Equal Contribution},
Zikang Wang$^{\clubsuit}$,
Ziqian Lin$^{\heartsuit}$,
Dongchen He$^{\heartsuit}$,\\
\bf Yuheng Shan$^{\diamondsuit}$,
Yanruisheng Shao$^{\heartsuit}$,
Jiayi Li$^{\heartsuit}$,
Xiangyu Shi$^{\heartsuit}$, 
Jiuming Wang$^{\heartsuit}$,\\
\bf Yanyu Chen$^{\heartsuit}$,
Yimin Fan$^{\heartsuit}$,
Han Li$^{\heartsuit}$,
Yu Li$^{\heartsuit}$\\
$^{\heartsuit}$ The Chinese University of Hong Kong,\\
$^{\spadesuit}$ Hong Kong University of Science and Technology (Guangzhou), \\
$^{\clubsuit}$ The Hong Kong Polytechnic University,
$^{\diamondsuit}$ National University of Singapore \\
}

\begin{document}
\maketitle
\begin{abstract}

Inverse Protein Folding (IPF) is a critical subtask in the field of protein design, aiming to engineer amino acid sequences capable of folding correctly into a specified three-dimensional (3D) conformation. Although substantial progress has been achieved in recent years, existing methods generally rely on either backbone coordinates or molecular surface features alone, which restricts their ability to fully capture the complex chemical and geometric constraints necessary for precise sequence prediction. To address this limitation, we present \textbf{DS-ProGen}, a dual-structure deep language model for functional protein design, which integrates both backbone geometry and surface-level representations. By incorporating backbone coordinates as well as surface chemical and geometric descriptors into a next-amino-acid prediction paradigm, DS-ProGen is able to generate functionally relevant and structurally stable sequences while satisfying both global and local conformational constraints. On the PRIDE dataset\footnote{A benchmark for structure-guided protein design evaluation, \url{https://github.com/chq1155/PRIDE_Benchmark_ProteinDesign}}, DS-ProGen attains the current state-of-the-art recovery rate of \textbf{61.47\%}, demonstrating the synergistic advantage of multi-modal structural encoding in protein design. Furthermore, DS-ProGen excels in predicting interactions with a variety of biological partners, including ligands, ions, and RNA, confirming its robust functional retention capabilities.
\end{abstract}

\section{Introduction}

Inverse Protein Folding remains a pivotal topic in molecular biology and protein engineering, aiming to design protein sequences with targeted functionalities for given three-dimensional structures~\cite{yue1992inverse, zhou2023prorefiner, gao2023proteininvbench}. Recently, significant strides have been made in text processing through advancements in Natural Language Processing (NLP) and Large Language Models (LLMs)~\cite{gu2021domain, achiam2023gpt, jiangetal2023cognitive, liu2024deepseek}. These methodologies have been successfully applied to the ``language'' of biology~\cite{ferruz2022protgpt2, madani2023large, jiang2025biological, nijkamp2023progen2, wang2025large, jiang2025benchmarking}, such as sequences and structural bioinformatics, offering fresh perspectives and technological foundations for prediction of sequence-structure relationships~\cite{xiao2024proteingpt, heinzinger2024bilingual}.

\begin{figure}[!t]
	\setlength{\belowcaptionskip}{-8pt}
	\centering
	\includegraphics[width=1\linewidth]{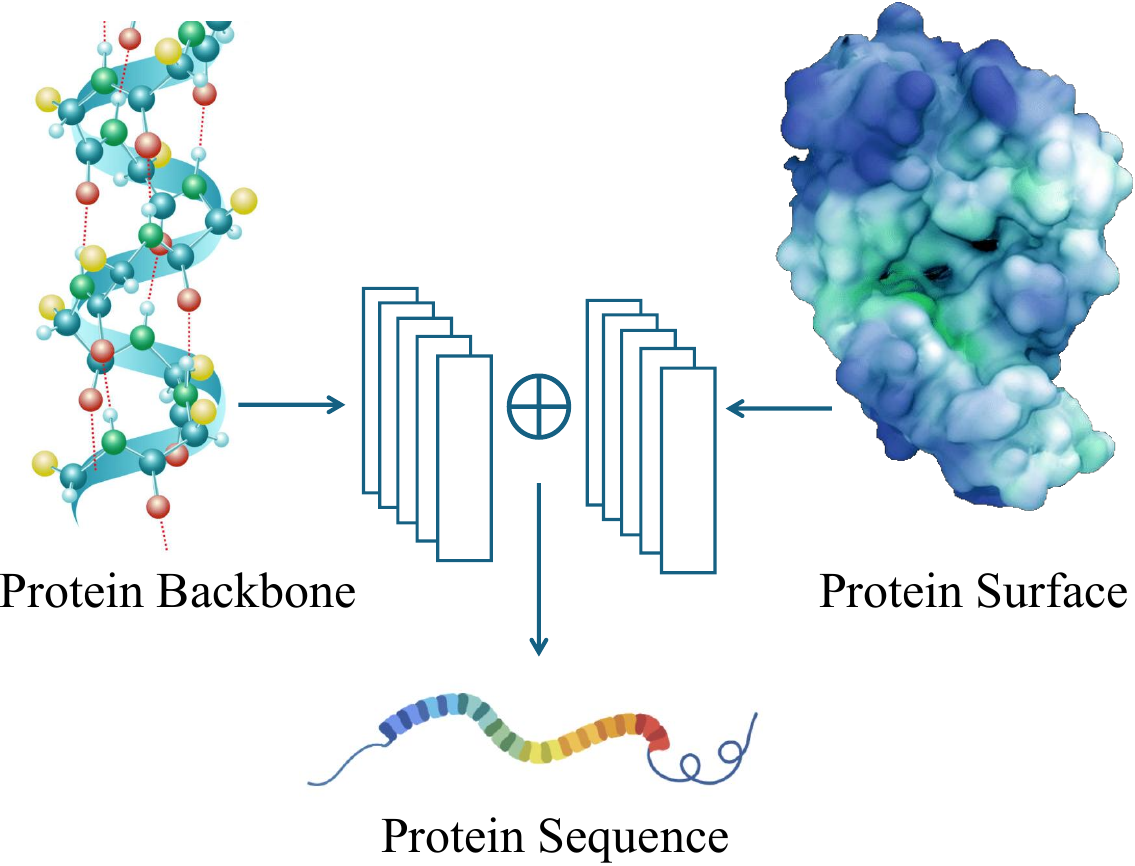}
	\caption{Design protein sequences based on protein structure (backbone) and surface features.} 
	\label{fig:example}
\end{figure}

Despite notable progress in applying language model approaches to protein science~\cite{ferruz2022protgpt2, madani2023large, nijkamp2023progen2}, most existing models for inverse folding rely heavily on singular structural representations either utilizing backbone geometric information, like skeletal coordinates~\cite{dauparas2022robust, jendrusch2021alphadesign}, or focusing solely on molecular surfaces~\cite{songsurfpro}. While backbone coordinates can reflect the intrinsic topology of the entire protein structure, they fail to capture crucial chemical features on the exterior surface that interact with the environment or ligands~\cite{emonts2023overview, cao2022design}. Conversely, surface information tends to overlook the influence of backbone conformations on the overall stability and functionality of proteins~\cite{songsurfpro}. Therefore, a significant challenge remains in achieving comprehensive modeling of both the ``internal-external'' structures in inverse folding tasks.


Our study introduces the DS-ProGen (Dual-Structure Protein Generator) model, which uniquely encodes both protein backbone and molecular surface features (Figure~\ref{fig:example}). Unlike unimodal models, DS-ProGen uses a backbone encoder to capture main chain coordinates, dihedral angles, and local geometric vectors, crucial for understanding internal structures. The surface encoder processes atomic types, curvature distributions, and chemical environments, focusing on micro-features at active sites and binding interfaces. These dual streams of information feed into a multimodal space and a multi-layer decoder that applies autoregressive techniques in sequence modeling. This approach allows for a holistic consideration of both geometric layout and surface chemical features when predicting amino acids. DS-ProGen outperforms traditional methods that rely solely on backbone coordinates or surface features, offering a richer structural and chemical analysis during inverse folding. Tests confirm its enhanced sequence generation capabilities and high functional accuracy in scenarios like ligand binding and complex molecular interactions.

The main contributions of our paper are: (1) Introducing the first multimodal protein language model for inverse folding, capturing backbone and surface chemical details. (2) Using geometric vectors with autoregressive Transformer decoding to improve sequence accuracy and structural fidelity. (3) Achieving higher sequence recovery rates on benchmarks and producing proteins viable for ligand and ion interactions. (4) Demonstrating enhancements in structural reconstruction and functional retention, with significant implications for drug discovery and synthetic biology.

\section{Related Works}

\subsection{Language Models for Scientific Applications}
Transformer-based language models \cite{devlin2019bert,radford2018improving} have been adapted to scientific domains, which often require structured representations and domain-specific vocabularies. Recent works like NatureLM \cite{xia2025naturelm} and UniGenX \cite{zhang2025unigenx} show that combining language models with scientific data accelerates tasks such as molecule synthesis and discovery.

\subsection{Protein Language Models}
Protein sequences mirror natural language properties, making them suitable for language modeling. Masked models like ESM \cite{lin2023evolutionary, hayes2025simulating} capture structural and functional features through large-scale sequence pretraining, enabling tasks like function prediction. Autoregressive models such as ProGen2 \cite{nijkamp2023progen2} and ProtGPT2 \cite{ferruz2022protgpt2} generate realistic proteins and support zero-shot fitness prediction, demonstrating strong design potential.

\subsection{Inverse Protein Folding}
Inverse protein folding \cite{yue1992inverse} aims to generate amino acid sequences for a desired target structure. Classical approaches like Rosetta \cite{das2008macromolecular} can be computationally intensive, while recent deep learning methods learn structure-conditioned sequence distributions. ProteinMPNN \cite{dauparas2022robust} and ESM-IF \cite{hsu2022learning} primarily rely on backbone geometry for strong sequence recovery. Other methods, such as PiFold \cite{gao2022pifold} and SurfPro \cite{songsurfpro}, encode local geometry or surface annotations. These approaches reveal a trade-off between internal structure capture and external interface modeling, and most still underutilize natural sequence evolutionary priors.

\begin{figure*}[htbp]
    \centering
    \includegraphics[width=\textwidth]{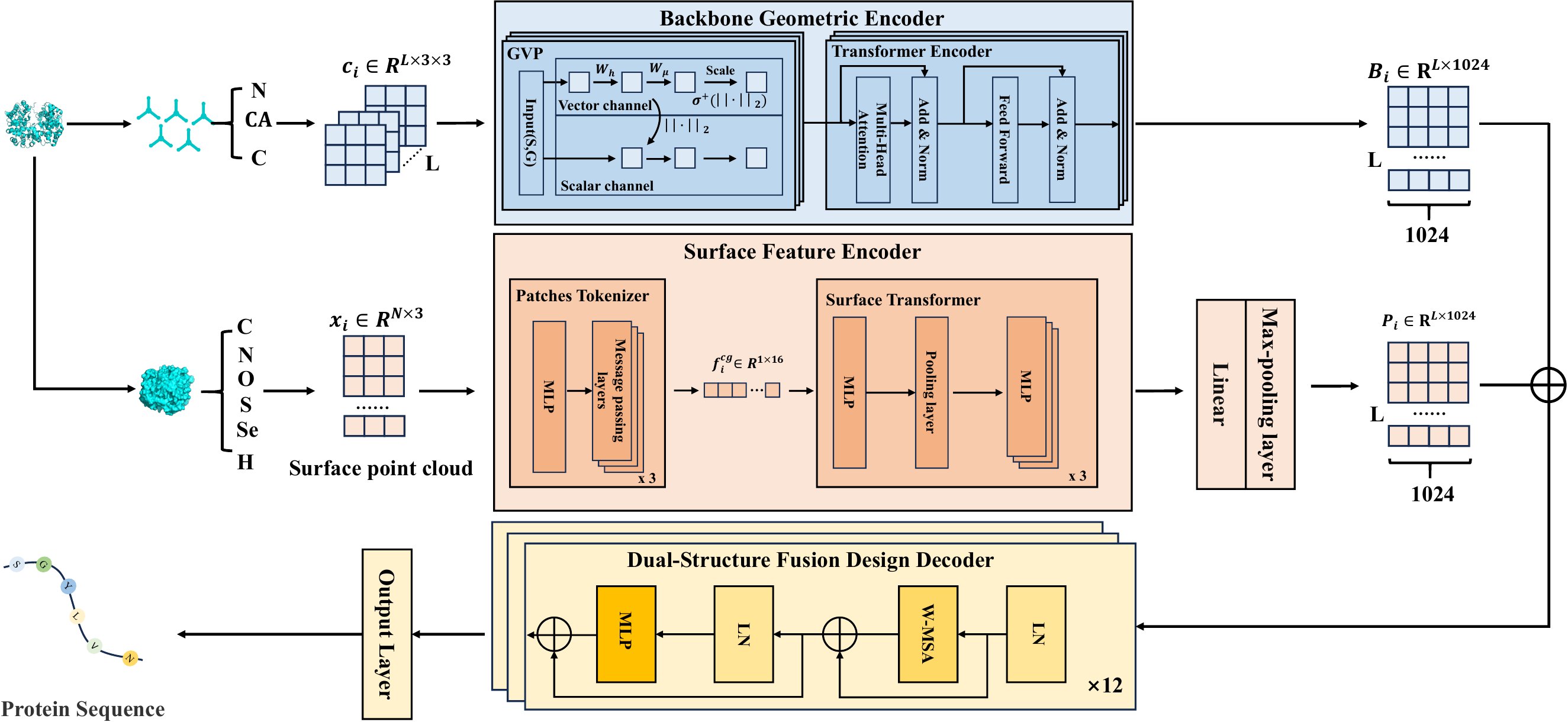} 
    \caption{DS-ProGen integrates backbone and surface structural information to enable functional protein design. The backbone geometric encoder extracts geometric features from the N, C, and C$_\alpha$ atoms of protein structures. The surface feature encoder processes atomic types, surface points, and curvature features. The backbone and surface embeddings are projected into a unified hidden space and combined to form a single representation with the target sequence length. Then, a dual-structure fusion design decoder auto-regressively generates/designs the protein sequence conditioned on the structural context.}
    \label{fig:framework}
\end{figure*}

\section{Methods}

To address these limitations, we propose DS-ProGen, a dual-structure deep language model for functional protein design that integrates both backbone and surface information to comprehensively characterize protein structures. DS-ProGen consists of a dual-branch encoder that extracts geometric and biochemical features from backbone coordinates and molecular surfaces, respectively, and a dual-structure fusion design
decoder. An overview of the architecture is illustrated in Figure~\ref{fig:framework}.

\subsection{Backbone Geometric Encoder}

We propose a backbone geometric encoder to extract the structural features of a protein based on the 3D coordinates of its backbone atoms.

To construct the input embeddings, we first extract the backbone atoms, namely, nitrogen (N), carbon (C), and alpha carbon (C$_\alpha$), from each residue. Their 3D coordinates are concatenated into a tensor $c_i \in \mathbb{R}^{L \times 3 \times 3}$, where $L$ is the sequence length. This coarse-grained representation retains the primary chain geometry while discarding side chains, providing a compact but informative description of protein structure.

Each residue is represented by a combination of scalar and vector features: the scalar feature $\mathbf{s} \in \mathbb{R}^{d_s}$ includes dihedral angles and pairwise distances, while the vector feature $\mathbf{V} \in \mathbb{R}^{d_v \times 3}$ encodes local spatial orientations. These features are updated using Geometric Vector Perceptron (GVP) layers~\cite{jing2020learning}, which are designed to be rotation-equivariant for vector channels and rotation-invariant for scalar channels. In our implementation, we use 4-layer GVP, each with dimensionality $d_s = 128$ and $d_v = 16$.

The update rules within each GVP layer are given by:
\begin{equation}
\begin{aligned}
\mathbf{V}' &= \sigma^+(\mathbf{v}_\mu) \odot \mathbf{V}_\mu, \\
\mathbf{s}' &= \sigma\left( W_m [\mathbf{s}; \mathbf{s}_h] + \mathbf{b} \right), \\
\mathbf{V}_h &= W_h \mathbf{V}, \quad
\mathbf{V}_\mu = W_\mu \mathbf{V}_h, \\
\mathbf{s}_h &= \| \mathbf{V}_h \|_2, \quad
\mathbf{v}_\mu = \| \mathbf{V}_\mu \|_2,
\end{aligned}
\end{equation}
where $W_h$, $W_\mu$, and $W_m$ are trainable weight matrices, and $\sigma$, $\sigma^+$ denote activation functions for scalar and vector channels, respectively.

After the GVP layers, a Transformer encoder~\cite{vaswani2017attention} with 8 layers is applied to capture relational dependencies among residues via multi-head self-attention. The final output is a sequence of embeddings $B \in \mathbb{R}^{L \times h_s}$, where $h_s = 1024$ matches the hidden state dimension of the sequence decoder.


\subsection{Surface Feature Encoder}

In addition to backbone structural features, the surface information of a protein provides complementary insights into its three-dimensional geometry. To leverage this additional structural perspective, we design a dedicated surface encoder capable of extracting and encoding surface information.

\subsubsection*{Surface Features}

Our representation of the protein surface incorporates three major components: atomic features, chemical and geometric properties of the surface points, and local curvature information.

We extract both the atomic coordinates and atom types. Formally, for each protein chain, we denote the atom coordinates as \(\{ a_i \}_{i=1}^M\) and atom types as one-hot encodings \(\{ t_i \}_{i=1}^{M}\), where:
\begin{equation}
a_i \in \mathbb{R}^{1\times 3}, \ t_i \in \mathbb{R}^{1\times 6}.
\end{equation}
Here, \(M\) is the total number of atoms. To simplify modeling and align with existing atom embeddings, we restrict our focus to six common element types (C, N, O, S, Se, H) and discard proteins containing other atom types.

To describe the external geometry of proteins, we construct surface point clouds. We apply a smooth distance function~\cite{blinn1982generalization} and van der Waals radii~\cite{batsanov2001van} to generate accurate surface representations. The resulting surface points are denoted as:

\begin{equation}
\{ x_i \}_{i=1}^N, \ x_i \in \mathbb{R}^{1\times 3},
\end{equation}

where \(N\) is the total number of points sampled from the surface. To make the batched training available, as previous literature\cite{yuan2023proteinmae} had done, we set the limitation of the point as 8,192. If the surface of the protein has more than 8,192 points, we will randomly choose 8,192 points as the input of surface information. If the surface data includes fewer than 8,192 points, we would randomly choose the points in the surface data above and then use them to pad the data to 8,192 points.

To enrich the surface points with local geometric and chemical context, we further compute detailed features around each surface point. For geometry, we derive normal vectors \(n_i \in \mathbb{R}^{1\times 3}\) and multiscale mean and Gaussian curvatures \(u_i\) across five radii (1Å, 2Å, 3Å, 5Å, 10Å):

\begin{equation}
\{ u_i \}_{i=1}^N, \ u_i \in \mathbb{R}^{1\times 10}.
\end{equation}

For chemical context, we consider the 16 nearest neighboring atoms around each surface point, following previous work~\cite{yuan2023proteinmae,sverrisson2021fast}, which provides a good trade-off between capturing sufficient local chemical information and maintaining computational efficiency. The distance between the \(i\)-th surface point and its \(j\)-th neighboring atom is:
\begin{equation}
d_{ij} = \| a_j - x_i \|,
\end{equation}
And the complete neighborhood feature is:
\begin{equation}
\left\{ (t_j, d_{ij}) \mid a_j \in N_i \right\} \in \mathbb{R}^{16 \times 7},
\end{equation}
where \(N_i\) denotes the set of 16 nearest atoms to \(x_i\), and \(t_j\) is the one-hot atom type encoding.

\subsubsection*{Patches Tokenizer}

Handling full surface point clouds is computationally prohibitive. Inspired by vision transformer (ViT) models~\cite{tolstikhin2021mlp}, we segment the surface into local patches. We first select \(g\) center points via farthest point sampling (FPS)~\cite{eldar1997farthest}. For each selected center \(c_i\), we gather a local patch using \(K\)-nearest neighbors (KNN)~\cite{zhang2016introduction}:

\begin{equation}
\{ P_i \}_{i=1}^g = \text{KNN}(\{ c_i \}_{i=1}^g, \{ x_j \}_{j=1}^N) \in \mathbb{R}^{g \times K \times 3},
\end{equation}

where each patch \(P_i\) contains \(K\) surface points, centered around \(c_i\).

Each patch is further enriched by embedding the one-hot atom types. An MLP followed by three message-passing layers encodes the atom features, producing a feature vector:

\begin{equation}
f_i^c \in \mathbb{R}^{1\times 6}.
\end{equation}

To integrate geometric context, we concatenate the chemical embedding \(f_i^c\) with the curvature vector \(u_i\), yielding:

\begin{equation}
f_i^{c_g} = [f_i^c; u_i] \in \mathbb{R}^{1\times 16}.
\end{equation}

This combined feature is updated through an MLP:

\begin{equation}
f_i^{c_g(l+1)} = f_i^{c_g(l)} + \text{MLP}(f_i^{c_g(l)}),
\end{equation}
starting from \(f_i^{c_g(0)} = f_i^{c_g}\), \(f_i \in \mathbb{R}\).

\subsubsection*{Surface Transformer}

The final features of all \(g\) patches are passed through a Transformer block~\cite{vaswani2017attention} to model global interactions across the surface. Following feature aggregation, we project the \(g \times d\) feature matrix into an \(h_g\times h_s\) space via two consecutive linear layers, where $h_g$ is the dimension of surface atomic features and $h_s$ is the hidden state dimension of the sequence decoder layers. A max-pooling operation along the surface points reduces the dimensionality, yielding \(S \in \mathbb{R}^{L\times h_s}\), where \(L\) denotes the target sequence length.

\subsection{Dual-Structure Fusion Design Decoder}

Our model adopts a Transformer decoder architecture to model the conditional generation of amino acid sequences~\cite{vaswani2017attention, nijkamp2023progen2}. In this framework, the sequence modeling task is formulated in an auto-regressive manner, where the joint probability of a sequence $\mathbf{x} = \{x_1, x_2, \ldots, x_T\}$ is factorized as a product of conditional probabilities:
\begin{equation}
P(\mathbf{x}) = \prod_{t=1}^{T} P(x_t \mid x_1, x_2, \ldots, x_{t-1}).
\end{equation}
At each time step, the decoder predicts the next token based solely on the preceding tokens, ensuring causality in generation.

To provide the model with structural context, we first integrate the backbone and surface representations. Specifically, we compute the combined structural embedding as:
\begin{equation}
R = B + S,
\end{equation}
where $B$ and $S$ are the backbone and surface feature embeddings, both projected into the same hidden dimension $h_s$.

The sequence decoder input is then constructed by prepending the combined structural embeddings $R$ as condition to the beginning of the sequence input embeddings. Formally, the decoder input matrix is:
\begin{equation}
\mathbf{X} = [R; \mathbf{E_t}],
\end{equation}
where $R \in \mathbb{R}^{L \times h_s}$ are the structural embeddings and $\mathbf{E_t} \in \mathbb{R}^{t \times h_s}$ are the sequence token embeddings.

For an input embedding matrix $\mathbf{X} \in \mathbb{R}^{t \times h_s}$, the Transformer decoder processes the sequence through stacked self-attention and feed-forward layers. The output hidden states $\mathbf{H} \in \mathbb{R}^{t \times h_s}$ are then projected by a linear layer to obtain the logits over the amino acid vocabulary: \begin{equation} \mathbf{Z} = \mathbf{H} \mathbf{W}_{\text{out}} + \mathbf{b}_{\text{out}}, \end{equation} where $\mathbf{W}_{\text{out}} \in \mathbb{R}^{h_s \times V}$ and $V$ is the vocabulary size. At each position, the next amino acid is predicted based on the logits.

Our implementation follows the GPT-2~\cite{radford2019language, nijkamp2023progen2} architecture, using stacked Transformer decoder layers to progressively refine the sequence representations and predict the next amino acid at each step.

\begin{table*}[t]
\centering
\begin{tabular}{lcccc}
\toprule
\textbf{Model} & \texttt{0<len<100 $\uparrow$} & \texttt{100$\leq$len<300 $\uparrow$} & \texttt{300$\leq$len<500 $\uparrow$} & \textbf{Overall $\uparrow$} \\
\midrule
ProteinMPNN  & 41.63 & 48.61 & 52.07 & 48.44 \\
PiFold       & 43.75 & 52.24 & 55.33 & 51.74 \\
ESM-IF        & 39.73 & 52.72 & 58.64 & 51.85 \\
\midrule
DS-ProGen (backbone-only) & 43.14 & 55.18 & 60.27 & 52.61 \\
DS-ProGen       & \textbf{63.50} & \textbf{64.46} & \textbf{63.15} & \textbf{61.47} \\
\bottomrule
\end{tabular}
\caption{The average recovery rate (\%) of DS-ProGen and baseline models on the PRIDE test set, grouped by sequence length. \textbf{Bold} indicates the best result in each column.}
\label{tab:Baseline Model}
\end{table*}

\section{Experiments}

\subsection{Experimental Setup}

\subsubsection*{Datasets}

Our training procedure is divided into a pretraining stage and a fine-tuning stage. Pretraining is motivated by the proven scaling laws of auto-regressive models, where performance systematically improves with larger model sizes and larger training datasets. Furthermore, prior work \cite{lin2023evolutionary} has demonstrated that pretraining significantly enhances the generalization ability of protein sequence models.


For the pretraining phase, we construct two datasets:
For backbone-only models, we gather over 40 million structure-sequence pairs from the FoldComp-compressed \cite{kim2023foldcomp} AlphaFoldDB Cluster Representatives \cite{jumper2021highly} and the high-quality ESMAtlas datasets \cite{lin2023evolutionary}, including only proteins with fewer than 512 residues.
For models learning from both backbone and surface, we select roughly 80,000 proteins from the same collection, focusing on those with valid surface data. This number balances the complexity of feature extraction, computational efficiency, and model performance. Our results confirm that using these 80,000 surface-annotated proteins achieves strong outcomes with manageable computational demands (Table~\ref{tab:ablation_pretraining}).



For fine-tuning and evaluation, We use the PRIDE benchmark \cite{chq1155pride2024} with 32,389 training proteins from CATH4.3 and 504 test proteins from CAMEO, ensuring structural diversity and difficulty. Details of the PRIDE benchmark can be found in Appendix~\ref{bench}

During fine-tuning and evaluation, the model with the surface encoder was trained only on samples with available surface information, or it just used the backbone information. The backbone-only model and baselines were trained on the full dataset without such restrictions.

During inference, we perform sampling to generate sequences. We use a temperature of 0.1 and apply top-$k$ sampling with $k=10$ to control the diversity of the generated sequences while maintaining high fidelity to the predicted distribution.

\subsubsection*{Implementation Details}

Before training, we initialize the model parameters as follows: for the sequence decoder, we load the pretrained weights from ProGen2-small~\cite{nijkamp2023progen2}; for the backbone encoder, we adopt the GVP and Transformer encoder parameters from ESM-IF~\cite{hsu2022learning}. During training, the backbone encoder is frozen to preserve its learned structural representations.

The overall model contains approximately 300M parameters. For the pretraining stage, we use a dataset of either 40 million or 80 thousand structure-sequence pairs. The model is trained for 1 epoch with a batch size of 16 using a single NVIDIA A100 GPU (80 GB memory) under FP32 precision. Training takes approximately 120 GPU-hours for the 40M dataset and 7 GPU-hours for the 80K dataset. The learning rate is set to $1 \times 10^{-4}$ with the Adam optimizer, using a 5\% warm-up phase followed by cosine annealing decay to zero.


For the fine-tuning stage, the model is trained for 2 epochs with a batch size of 8. The learning rate during fine-tuning is reduced to $2 \times 10^{-5}$, also using the Adam optimizer with cosine decay.

\begin{table*}[t]
\centering
\begin{tabular}{lcc|cc}
\toprule
\textbf{Model} & \multicolumn{2}{c|}{\textbf{TM-Score (\%) $\uparrow$}} & \multicolumn{2}{c}{\textbf{RMSD (Å) $\downarrow$}} \\
\cmidrule(lr){2-3} \cmidrule(lr){4-5}
 & Mean & Median & Mean & Median \\
\midrule
ProteinMPNN  & 87.58 & 94.72 & 1.935 & 0.7477 \\
PiFold       & \textbf{88.43} & 95.11 & 1.859 & 0.6448 \\
ESM-IF       & 88.38 & \textbf{95.58} & 1.652 & 0.5811 \\
\midrule
DS-ProGen (backbone-only) & 86.11 & 94.03 & 1.780 & 0.6087 \\
DS-ProGen     & 87.30 & 94.88 & \textbf{1.401} & \textbf{0.5575} \\
\bottomrule
\end{tabular}
\caption{TM-Score and RMSD measure between AlphaFold3-predicted structures from model-generated sequences and target structures, PRIDE test set. \textbf{Bold} indicates the best result in each column.}
\label{tab:tmscore_rmsd}
\end{table*}

\begin{figure*}[t]
    \centering
    \includegraphics[width=\linewidth]{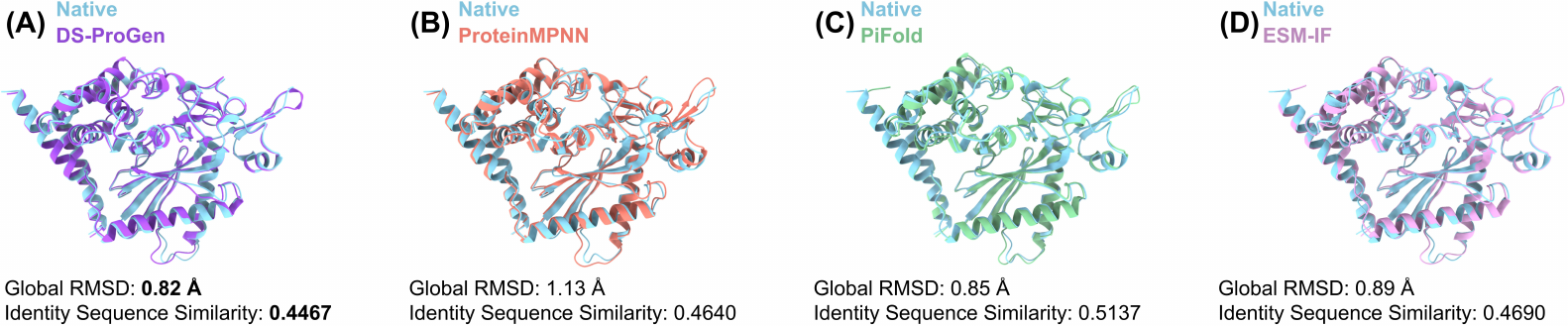}
    \caption{Design cases for DS-ProGen and baselines (Length is 403). A lower \textbf{RMSD} indicates greater structural similarity, while a lower \textbf{Identity Sequence Similarity} suggests stronger diversity in the designed sequences.}
    \label{fig:ss}
\end{figure*}

\subsection{Evaluation Metrics}
To evaluate the accuracy of sequence prediction, following previous research of inverse folding~\cite{jing2020learning, gao2024proteininvbench}, we use the \textbf{recovery rate} as the primary metric. The recovery rate measures the proportion of residues where the predicted amino acid matches the ground-truth amino acid at the same position.

Formally, given a ground-truth sequence $\mathbf{y} = (y_1, y_2, \ldots, y_L)$ and a predicted sequence $\hat{\mathbf{y}} = (\hat{y}_1, \hat{y}_2, \ldots, \hat{y}_L)$ of length $L$, the recovery rate $R$ is defined as:

\begin{equation} R = \frac{1}{L} \sum_{i=1}^{L} \mathbb{I}(y_i = \hat{y}_i), \end{equation}

where $\mathbb{I}(\cdot)$ is the indicator function, which equals $1$ if the condition inside is true and $0$ otherwise.

In addition to the recovery rate, to assess whether the generated sequences can correctly refold into the desired target structures, we further evaluate the predicted structures using AlphaFold3\cite{Abramson2024}. Specifically, for each generated sequence, we predict its three-dimensional structure and compare it with the native target structure using two metrics: TM-Score \cite{zhang2004scoring} and Root-Mean-Square Deviation (RMSD). Higher TM-Score and lower RMSD indicate better structural quality.

\subsection{Baselines}

We compare our method against three models: ProteinMPNN \cite{dauparas2022robust}, PiFold \cite{gao2022pifold}, and ESM-IF \cite{hsu2022learning}.
ProteinMPNN and PiFold are retrained on the PRIDE benchmark training set to ensure a fair comparison.
For ESM-IF, we directly evaluate the released pretrained model without further fine-tuning.
All baselines are based on their official implementations, with only minor modifications for environment compatibility.

\begin{figure*}[t]
    \centering
    \includegraphics[width=0.95\textwidth]{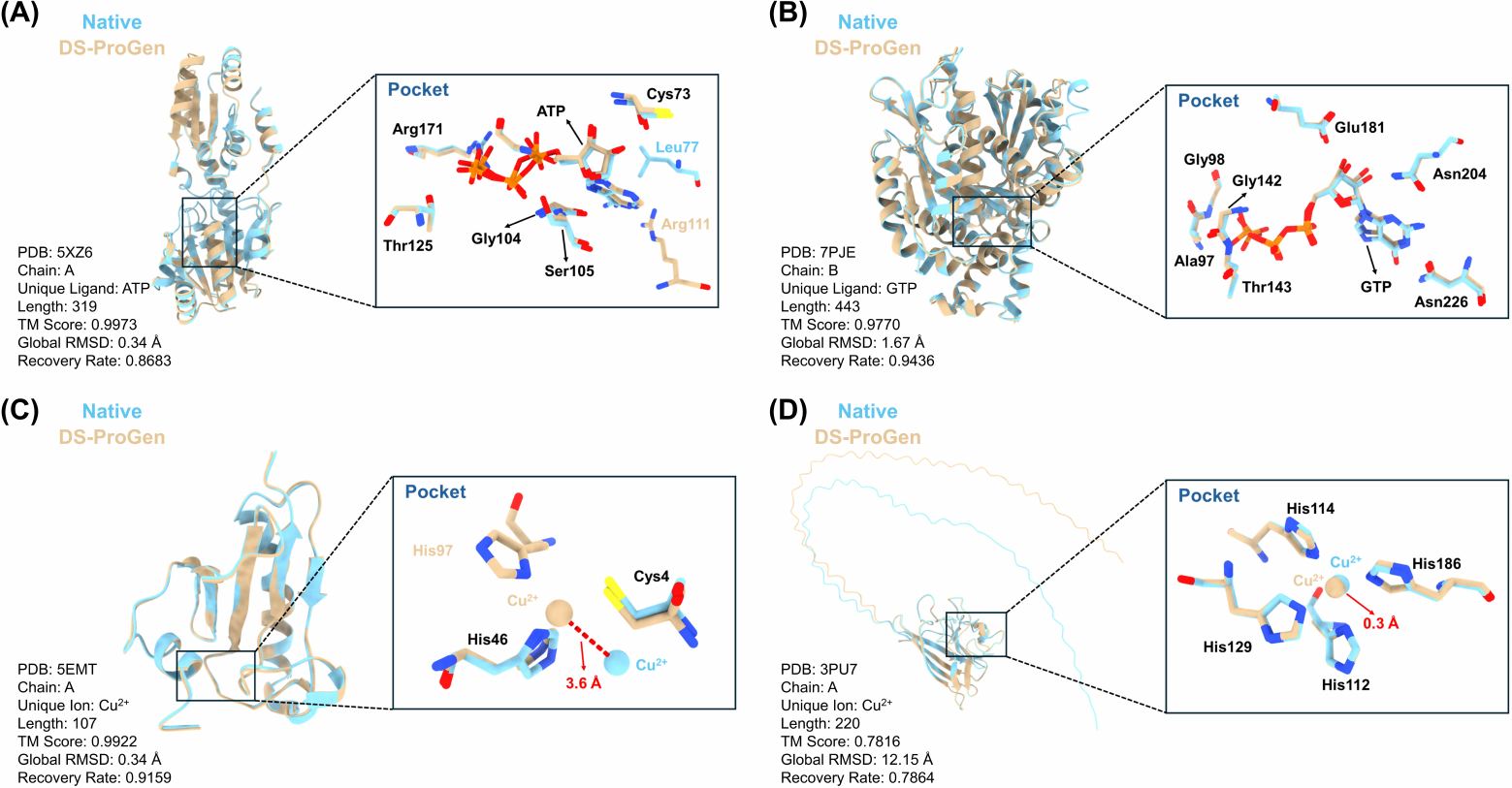}
    \caption{Structural alignment between ground truth structures (blue) and predicted structures (light brown) folded from protein sequences generated by DS-ProGen, visualized across four representative cases. Each panel highlights the global fold alignment and a zoomed-in view of the functional binding pocket, showing key residues together with the corresponding ligand or ion (e.g., ATP, GTP, \(\mathrm{Cu}^{2+}\)). The predicted sequences lead to highly accurate structural models, with high TM-scores, low RMSDs, and strong recovery rates, demonstrating DS-ProGen’s ability to preserve both global topology and local biochemical specificity. }
    \label{casestudy}
\end{figure*}

\subsection{Main Results}
As shown in Table~\ref{tab:Baseline Model}, DS-ProGen achieves the highest overall recovery rate of \textbf{61.47\%} on the PRIDE test set, consistently outperforming all baselines across sequence length ranges. The improvement is particularly notable on short sequences (\texttt{len<100}), where DS-ProGen reaches \textbf{63.50\%}, nearly 20 percentage points higher than the second-best model PiFold (43.75\%). This highlights the value of surface features in low-resolution backbone graphs, where node sparsity limits the capacity of graph-based encoders to extract representations.

Moreover, as shown in Figure~\ref{fig:recovery_rate_violin} and Table~\ref{tab:Baseline Model}, DS-ProGen not only achieves a higher median recovery rate but also exhibits a narrower distribution across test samples, suggesting greater robustness and reduced variance compared to the baselines.

Regarding structural fidelity, Table~\ref{tab:tmscore_rmsd} shows that while DS-ProGen achieves the lowest RMSD among all models, its TM-Score is marginally lower than those of PiFold and ESM-IF. We note that the structures used for evaluation are predicted by AlphaFold3, which, despite its high accuracy, still introduces prediction noise. Therefore, differences in TM-Score or RMSD do not definitively indicate superiority or inferiority in structural realism or design quality of the generated sequences.


Despite achieving a high recovery rate and structural similarity, the designed sequences exhibit only 61.9\% sequence similarity (Figure~\ref{fig:seq_sim}), indicating significant diversity. The overall trends are encouraging. DS-ProGen's performance aligns closely with the top-performing baselines, and the uniformly low RMSD values across the test set suggest our model reliably generates sequences that fold into physically plausible structures. This underscores the benefit of incorporating surface information, which not only boosts sequence recovery but also maintains or enhances structural compatibility, especially regarding local precision and foldability.

\subsection{Ablation Study}

To systematically assess the contributions of different model components, we conduct two sets of ablation studies within the DS-ProGen framework:
(1) encoder design, and 
(2) pretraining strategy.

\vspace{0.5em}
\noindent
\textbf{Encoder Design.}
We evaluate the impact of using the backbone encoder (b\_enc) and surface encoder (s\_enc). 
The results are summarized in Table~\ref{tab:ablation_study}.

\begin{table}[h]\small
\centering
\begin{tabular}{lc}
\toprule
Methods & Recovery Rate (\%) \\
\midrule
DS-ProGen             & 61.47 \\
w/o s\_enc           & 50.71 \\
w/o b\_enc           & 31.06 \\
\bottomrule
\end{tabular}
\caption{
Ablation study on the encoder design of DS-ProGen. 
All models are pretrained on the same 80k dataset with available surface information. 
Removing either the backbone encoder (b\_enc) or the surface encoder (s\_enc) results in performance degradation, highlighting their complementary contributions.
}
\label{tab:ablation_study}
\end{table}

As shown in Table~\ref{tab:ablation_study}, jointly using both the backbone and surface encoders yields the highest recovery rate (61.47\%). 
Removing the surface encoder (w/o s\_enc) leads to a moderate performance drop, while removing the backbone encoder (w/o b\_enc) causes a significant degradation. 
This indicates that backbone geometry provides strong global structure cues, while surface information captures important local features, and the combination of both is critical for accurate sequence design.

\vspace{1em}
\noindent
\textbf{Pretraining Strategy.}
We further examine the importance of pretraining by comparing models trained with/ and w/o initialization from large structure-sequence pairs.
The results are presented in Table~\ref{tab:ablation_pretraining}.

\begin{table}[h]\small
\centering
\begin{tabular}{lc}
\toprule
Methods & Recovery Rate (\%) \\
\midrule
DS-ProGen (backbone-only)   & 52.61 \\
w/o pretraining            & 33.62 \\
\bottomrule
\end{tabular}
\caption{
Ablation study on the effect of pretraining in DS-ProGen. 
Pretraining on 4M structure-sequence pairs significantly improves model performance compared to training from scratch, demonstrating its critical role in enhancing generalization and convergence.
}
\label{tab:ablation_pretraining}
\end{table}

As shown in Table~\ref{tab:ablation_pretraining}, pretraining brings a substantial gain of nearly 19\% in recovery rate (52.61\% vs. 33.62\%). 
This highlights the necessity of pretraining for establishing strong structure-aware sequence priors, which significantly accelerates convergence and boosts generalization.

\subsection{Downstream Tasks: Ligand and Ion-Protein Interaction Design}
Beyond the standard inverse folding task, our further expectation is to design functional proteins with practical utility. Specifically, given the structure of a protein known to bind with a particular ligand or ion, we used DS-ProGen to generate compatible amino acid sequences. Molecular recognition between proteins and ligands drives many biological processes~\cite{janin2008protein}, while ion-protein interactions are always playing critical roles in biochemical processes~\cite{roberts2015specific}. To evaluate DS-ProGen in these contexts, we applied it to design sequences for both ligand- and ion-binding proteins, then predicted their 3D structures using AlphaFold3 and employed UCSF ChimeraX~\cite{https://doi.org/10.1002/pro.3943} for visualizing structural similarity between the ground truth and the predicted structures. In both tasks, as illustrated in Figure~\ref{casestudy}, the generated sequences consistently yield highly accurate structural models, reflected by high TM-scores, low RMSDs and strong recovery rates, demonstrating that DS-ProGen can effectively design proteins with both strong binding capabilities and high structural fidelity to their targets. For comparison, we also performed structural alignment between native structures and predicted structures folded from the sequences generated by three different baselines (Figures~\ref{ProteinMPNN}, \ref{PiFold}, \ref{ESM-IF}).

\section{Conclusion and Outlook}



We introduce DS-ProGen, a dual-structure generative model for protein design that utilizes both backbone and surface features. By leveraging geometric and chemical cues, it surpasses single-modality baselines with a 61.47\% recovery rate on the PRIDE benchmark and excels in tasks like ligand and ion interaction design.

Future enhancements of DS-ProGen may include applications like RNA inverse folding and structure-guided ligand design. Its modular, autoregressive architecture supports the development of multi-task models on extensive biological datasets.



\section*{Limitations}
Despite the improvements of DS-ProGen, several limitations remain. Limited computational resources prevented large-scale training and extensive benchmarking, forcing us to initialize the model with pretrained weights rather than training from scratch. Additionally, the reliance on structural predictions from tools like AlphaFold, without accounting for potential prediction errors, may impact overall performance. Moreover, the fixed-size patching strategy for surface encoding, although efficient, can lead to loss of critical structural details, particularly in irregular or sparsely sampled regions.

In addition, we only use AI tools to polish the language of our paper.



\bibliography{custom}

\begin{thebibliography}{50}
\providecommand{\natexlab}[1]{#1}

\bibitem[{Abramson et~al.(2024)Abramson, Adler, Dunger, Evans, Green, Pritzel, Ronneberger, Willmore, Ballard, Bambrick, Bodenstein, Evans, Hung, O’Neill, Reiman, Tunyasuvunakool, Wu, Žemgulytė, Arvaniti, Beattie, Bertolli, Bridgland, Cherepanov, Congreve, Cowen-Rivers, Cowie, Figurnov, Fuchs, Gladman, Jain, Khan, Low, Perlin, Potapenko, Savy, Singh, Stecula, Thillaisundaram, Tong, Yakneen, Zhong, Zielinski, Žídek, Bapst, Kohli, Jaderberg, Hassabis, and Jumper}]{Abramson2024}
Josh Abramson, Jonas Adler, Jack Dunger, Richard Evans, Tim Green, Alexander Pritzel, Olaf Ronneberger, Lindsay Willmore, Andrew~J. Ballard, Joshua Bambrick, Sebastian~W. Bodenstein, David~A. Evans, Chia-Chun Hung, Michael O’Neill, David Reiman, Kathryn Tunyasuvunakool, Zachary Wu, Akvilė Žemgulytė, Eirini Arvaniti, and 29 others. 2024.
\newblock \href {https://doi.org/10.1038/s41586-024-07487-w} {Accurate structure prediction of biomolecular interactions with alphafold 3}.
\newblock \emph{Nature}, 630(8016):493–--500.

\bibitem[{Achiam et~al.(2023)Achiam, Adler, Agarwal, Ahmad, Akkaya, Aleman, Almeida, Altenschmidt, Altman, Anadkat et~al.}]{achiam2023gpt}
Josh Achiam, Steven Adler, Sandhini Agarwal, Lama Ahmad, Ilge Akkaya, Florencia~Leoni Aleman, Diogo Almeida, Janko Altenschmidt, Sam Altman, Shyamal Anadkat, and 1 others. 2023.
\newblock Gpt-4 technical report.
\newblock \emph{arXiv preprint arXiv:2303.08774}.

\bibitem[{Batsanov(2001)}]{batsanov2001van}
Stepan~S Batsanov. 2001.
\newblock Van der waals radii of elements.
\newblock \emph{Inorganic materials}, 37(9):871--885.

\bibitem[{Blinn(1982)}]{blinn1982generalization}
James~F Blinn. 1982.
\newblock A generalization of algebraic surface drawing.
\newblock \emph{ACM transactions on graphics (TOG)}, 1(3):235--256.

\bibitem[{Camacho et~al.(2009)Camacho, Coulouris, Avagyan, Ma, Papadopoulos, Bealer, and Madden}]{camacho2009blast+}
Christiam Camacho, George Coulouris, Vahram Avagyan, Ning Ma, Jason Papadopoulos, Kevin Bealer, and Thomas~L Madden. 2009.
\newblock Blast+: architecture and applications.
\newblock \emph{BMC bioinformatics}, 10:1--9.

\bibitem[{Cao et~al.(2022)Cao, Coventry, Goreshnik, Huang, Sheffler, Park, Jude, Markovi{\'c}, Kadam, Verschueren et~al.}]{cao2022design}
Longxing Cao, Brian Coventry, Inna Goreshnik, Buwei Huang, William Sheffler, Joon~Sung Park, Kevin~M Jude, Iva Markovi{\'c}, Rameshwar~U Kadam, Koen~HG Verschueren, and 1 others. 2022.
\newblock Design of protein-binding proteins from the target structure alone.
\newblock \emph{Nature}, 605(7910):551--560.

\bibitem[{chq1155(2024)}]{chq1155pride2024}
chq1155. 2024.
\newblock Pride\_benchmark\_proteindesign.
\newblock \url{https://github.com/chq1155/PRIDE_Benchmark_ProteinDesign}.
\newblock Accessed: 2025-04-26.

\bibitem[{Das and Baker(2008)}]{das2008macromolecular}
Rhiju Das and David Baker. 2008.
\newblock Macromolecular modeling with rosetta.
\newblock \emph{Annu. Rev. Biochem.}, 77(1):363--382.

\bibitem[{Dauparas et~al.(2022)Dauparas, Anishchenko, Bennett, Bai, Ragotte, Milles, Wicky, Courbet, de~Haas, Bethel et~al.}]{dauparas2022robust}
Justas Dauparas, Ivan Anishchenko, Nathaniel Bennett, Hua Bai, Robert~J Ragotte, Lukas~F Milles, Basile~IM Wicky, Alexis Courbet, Rob~J de~Haas, Neville Bethel, and 1 others. 2022.
\newblock Robust deep learning--based protein sequence design using proteinmpnn.
\newblock \emph{Science}, 378(6615):49--56.

\bibitem[{Devlin et~al.(2019)Devlin, Chang, Lee, and Toutanova}]{devlin2019bert}
Jacob Devlin, Ming-Wei Chang, Kenton Lee, and Kristina Toutanova. 2019.
\newblock Bert: Pre-training of deep bidirectional transformers for language understanding.
\newblock In \emph{Proceedings of the 2019 conference of the North American chapter of the association for computational linguistics: human language technologies, volume 1 (long and short papers)}, pages 4171--4186.

\bibitem[{Eldar et~al.(1997)Eldar, Lindenbaum, Porat, and Zeevi}]{eldar1997farthest}
Yuval Eldar, Michael Lindenbaum, Moshe Porat, and Yehoshua~Y Zeevi. 1997.
\newblock The farthest point strategy for progressive image sampling.
\newblock \emph{IEEE transactions on image processing}, 6(9):1305--1315.

\bibitem[{Emonts and Buyel(2023)}]{emonts2023overview}
Jessica Emonts and Johannes~Felix Buyel. 2023.
\newblock An overview of descriptors to capture protein properties--tools and perspectives in the context of qsar modeling.
\newblock \emph{Computational and structural biotechnology journal}, 21:3234--3247.

\bibitem[{Ferruz et~al.(2022)Ferruz, Schmidt, and H{\"o}cker}]{ferruz2022protgpt2}
Noelia Ferruz, Steffen Schmidt, and Birte H{\"o}cker. 2022.
\newblock Protgpt2 is a deep unsupervised language model for protein design.
\newblock \emph{Nature communications}, 13(1):4348.

\bibitem[{Gao et~al.(2022)Gao, Tan, Chac{\'o}n, and Li}]{gao2022pifold}
Zhangyang Gao, Cheng Tan, Pablo Chac{\'o}n, and Stan~Z Li. 2022.
\newblock Pifold: Toward effective and efficient protein inverse folding.
\newblock \emph{arXiv preprint arXiv:2209.12643}.

\bibitem[{Gao et~al.(2023)Gao, Tan, Zhang, Chen, Wu, and Li}]{gao2023proteininvbench}
Zhangyang Gao, Cheng Tan, Yijie Zhang, Xingran Chen, Lirong Wu, and Stan~Z Li. 2023.
\newblock Proteininvbench: Benchmarking protein inverse folding on diverse tasks, models, and metrics.
\newblock \emph{Advances in Neural Information Processing Systems}, 36:68207--68220.

\bibitem[{Gao et~al.(2024)Gao, Tan, Zhang, Chen, Wu, and Li}]{gao2024proteininvbench}
Zhangyang Gao, Cheng Tan, Yijie Zhang, Xingran Chen, Lirong Wu, and Stan~Z Li. 2024.
\newblock Proteininvbench: Benchmarking protein inverse folding on diverse tasks, models, and metrics.
\newblock \emph{Advances in Neural Information Processing Systems}, 36.

\bibitem[{Gu et~al.(2021)Gu, Tinn, Cheng, Lucas, Usuyama, Liu, Naumann, Gao, and Poon}]{gu2021domain}
Yu~Gu, Robert Tinn, Hao Cheng, Michael Lucas, Naoto Usuyama, Xiaodong Liu, Tristan Naumann, Jianfeng Gao, and Hoifung Poon. 2021.
\newblock Domain-specific language model pretraining for biomedical natural language processing.
\newblock \emph{ACM Transactions on Computing for Healthcare (HEALTH)}, 3(1):1--23.

\bibitem[{Hayes et~al.(2025)Hayes, Rao, Akin, Sofroniew, Oktay, Lin, Verkuil, Tran, Deaton, Wiggert et~al.}]{hayes2025simulating}
Thomas Hayes, Roshan Rao, Halil Akin, Nicholas~J Sofroniew, Deniz Oktay, Zeming Lin, Robert Verkuil, Vincent~Q Tran, Jonathan Deaton, Marius Wiggert, and 1 others. 2025.
\newblock Simulating 500 million years of evolution with a language model.
\newblock \emph{Science}, page eads0018.

\bibitem[{Heinzinger et~al.(2024)Heinzinger, Weissenow, Sanchez, Henkel, Mirdita, Steinegger, and Rost}]{heinzinger2024bilingual}
Michael Heinzinger, Konstantin Weissenow, Joaquin~Gomez Sanchez, Adrian Henkel, Milot Mirdita, Martin Steinegger, and Burkhard Rost. 2024.
\newblock Bilingual language model for protein sequence and structure.
\newblock \emph{NAR Genomics and Bioinformatics}, 6(4):lqae150.

\bibitem[{Hsu et~al.(2022)Hsu, Verkuil, Liu, Lin, Hie, Sercu, Lerer, and Rives}]{hsu2022learning}
Chloe Hsu, Robert Verkuil, Jason Liu, Zeming Lin, Brian Hie, Tom Sercu, Adam Lerer, and Alexander Rives. 2022.
\newblock Learning inverse folding from millions of predicted structures.
\newblock In \emph{International conference on machine learning}, pages 8946--8970. PMLR.

\bibitem[{Janin et~al.(2008)Janin, Bahadur, and Chakrabarti}]{janin2008protein}
Jo{\"e}l Janin, Ranjit~P Bahadur, and Pinak Chakrabarti. 2008.
\newblock Protein--protein interaction and quaternary structure.
\newblock \emph{Quarterly reviews of biophysics}, 41(2):133--180.

\bibitem[{Jendrusch et~al.(2021)Jendrusch, Korbel, and Sadiq}]{jendrusch2021alphadesign}
Michael Jendrusch, Jan~O Korbel, and S~Kashif Sadiq. 2021.
\newblock Alphadesign: A de novo protein design framework based on alphafold.
\newblock \emph{Biorxiv}, pages 2021--10.

\bibitem[{Jiang et~al.(2025{\natexlab{a}})Jiang, Chen, Wang, He, Wei, Hong, Zong, Wang, Yu, Ma et~al.}]{jiang2025benchmarking}
Jiyue Jiang, Pengan Chen, Jiuming Wang, Dongchen He, Ziqin Wei, Liang Hong, Licheng Zong, Sheng Wang, Qinze Yu, Zixian Ma, and 1 others. 2025{\natexlab{a}}.
\newblock Benchmarking large language models on multiple tasks in bioinformatics nlp with prompting.
\newblock \emph{arXiv preprint arXiv:2503.04013}.

\bibitem[{Jiang et~al.(2023)Jiang, Wang, Li, Kong, and Wu}]{jiangetal2023cognitive}
Jiyue Jiang, Sheng Wang, Qintong Li, Lingpeng Kong, and Chuan Wu. 2023.
\newblock \href {https://doi.org/10.18653/v1/2023.acl-long.593} {A cognitive stimulation dialogue system with multi-source knowledge fusion for elders with cognitive impairment}.
\newblock In \emph{Proceedings of the 61st Annual Meeting of the Association for Computational Linguistics (Volume 1: Long Papers)}, pages 10628--10640, Toronto, Canada. Association for Computational Linguistics.

\bibitem[{Jiang et~al.(2025{\natexlab{b}})Jiang, Wang, Shan, Chai, Li, Ma, Zhang, and Li}]{jiang2025biological}
Jiyue Jiang, Zikang Wang, Yuheng Shan, Heyan Chai, Jiayi Li, Zixian Ma, Xinrui Zhang, and Yu~Li. 2025{\natexlab{b}}.
\newblock Biological sequence with language model prompting: A survey.
\newblock \emph{arXiv preprint arXiv:2503.04135}.

\bibitem[{Jing et~al.(2020)Jing, Eismann, Suriana, Townshend, and Dror}]{jing2020learning}
Bowen Jing, Stephan Eismann, Patricia Suriana, Raphael~JL Townshend, and Ron Dror. 2020.
\newblock Learning from protein structure with geometric vector perceptrons.
\newblock \emph{arXiv preprint arXiv:2009.01411}.

\bibitem[{Jumper et~al.(2021)Jumper, Evans, Pritzel, Green, Figurnov, Ronneberger, Tunyasuvunakool, Bates, {\v{Z}}{\'\i}dek, Potapenko et~al.}]{jumper2021highly}
John Jumper, Richard Evans, Alexander Pritzel, Tim Green, Michael Figurnov, Olaf Ronneberger, Kathryn Tunyasuvunakool, Russ Bates, Augustin {\v{Z}}{\'\i}dek, Anna Potapenko, and 1 others. 2021.
\newblock Highly accurate protein structure prediction with alphafold.
\newblock \emph{nature}, 596(7873):583--589.

\bibitem[{Kim et~al.(2023)Kim, Mirdita, and Steinegger}]{kim2023foldcomp}
Hyunbin Kim, Milot Mirdita, and Martin Steinegger. 2023.
\newblock Foldcomp: a library and format for compressing and indexing large protein structure sets.
\newblock \emph{Bioinformatics}, 39(4):btad153.

\bibitem[{Lin et~al.(2023)Lin, Akin, Rao, Hie, Zhu, Lu, Smetanin, Verkuil, Kabeli, Shmueli et~al.}]{lin2023evolutionary}
Zeming Lin, Halil Akin, Roshan Rao, Brian Hie, Zhongkai Zhu, Wenting Lu, Nikita Smetanin, Robert Verkuil, Ori Kabeli, Yaniv Shmueli, and 1 others. 2023.
\newblock Evolutionary-scale prediction of atomic-level protein structure with a language model.
\newblock \emph{Science}, 379(6637):1123--1130.

\bibitem[{Liu et~al.(2024)Liu, Feng, Xue, Wang, Wu, Lu, Zhao, Deng, Zhang, Ruan et~al.}]{liu2024deepseek}
Aixin Liu, Bei Feng, Bing Xue, Bingxuan Wang, Bochao Wu, Chengda Lu, Chenggang Zhao, Chengqi Deng, Chenyu Zhang, Chong Ruan, and 1 others. 2024.
\newblock Deepseek-v3 technical report.
\newblock \emph{arXiv preprint arXiv:2412.19437}.

\bibitem[{Madani et~al.(2023)Madani, Krause, Greene, Subramanian, Mohr, Holton, Olmos, Xiong, Sun, Socher et~al.}]{madani2023large}
Ali Madani, Ben Krause, Eric~R Greene, Subu Subramanian, Benjamin~P Mohr, James~M Holton, Jose~Luis Olmos, Caiming Xiong, Zachary~Z Sun, Richard Socher, and 1 others. 2023.
\newblock Large language models generate functional protein sequences across diverse families.
\newblock \emph{Nature Biotechnology}, 41(8):1099--1106.

\bibitem[{Nijkamp et~al.(2023)Nijkamp, Ruffolo, Weinstein, Naik, and Madani}]{nijkamp2023progen2}
Erik Nijkamp, Jeffrey~A Ruffolo, Eli~N Weinstein, Nikhil Naik, and Ali Madani. 2023.
\newblock Progen2: exploring the boundaries of protein language models.
\newblock \emph{Cell systems}, 14(11):968--978.

\bibitem[{Pettersen et~al.(2021)Pettersen, Goddard, Huang, Meng, Couch, Croll, Morris, and Ferrin}]{https://doi.org/10.1002/pro.3943}
Eric~F. Pettersen, Thomas~D. Goddard, Conrad~C. Huang, Elaine~C. Meng, Gregory~S. Couch, Tristan~I. Croll, John~H. Morris, and Thomas~E. Ferrin. 2021.
\newblock \href {https://doi.org/10.1002/pro.3943} {Ucsf chimerax: Structure visualization for researchers, educators, and developers}.
\newblock \emph{Protein Science}, 30(1):70--82.

\bibitem[{Radford et~al.(2018)Radford, Narasimhan, Salimans, Sutskever et~al.}]{radford2018improving}
Alec Radford, Karthik Narasimhan, Tim Salimans, Ilya Sutskever, and 1 others. 2018.
\newblock Improving language understanding by generative pre-training.

\bibitem[{Radford et~al.(2019)Radford, Wu, Child, Luan, Amodei, Sutskever et~al.}]{radford2019language}
Alec Radford, Jeffrey Wu, Rewon Child, David Luan, Dario Amodei, Ilya Sutskever, and 1 others. 2019.
\newblock Language models are unsupervised multitask learners.
\newblock \emph{OpenAI blog}, 1(8):9.

\bibitem[{Roberts et~al.(2015)Roberts, Keeling, Tracka, Van Der~Walle, Uddin, Warwicker, and Curtis}]{roberts2015specific}
Dorota Roberts, R~Keeling, M~Tracka, CF~Van Der~Walle, S~Uddin, Jim Warwicker, and R~Curtis. 2015.
\newblock Specific ion and buffer effects on protein--protein interactions of a monoclonal antibody.
\newblock \emph{Molecular pharmaceutics}, 12(1):179--193.

\bibitem[{Song et~al.()Song, Huang, Li, and Jin}]{songsurfpro}
Zhenqiao Song, Tinglin Huang, Lei Li, and Wengong Jin.
\newblock Surfpro: Functional protein design based on continuous surface.
\newblock In \emph{Forty-first International Conference on Machine Learning}.

\bibitem[{Sverrisson et~al.(2021)Sverrisson, Feydy, Correia, and Bronstein}]{sverrisson2021fast}
Freyr Sverrisson, Jean Feydy, Bruno~E Correia, and Michael~M Bronstein. 2021.
\newblock Fast end-to-end learning on protein surfaces.
\newblock In \emph{Proceedings of the IEEE/CVF Conference on Computer Vision and Pattern Recognition}, pages 15272--15281.

\bibitem[{Tolstikhin et~al.(2021)Tolstikhin, Houlsby, Kolesnikov, Beyer, Zhai, Unterthiner, Yung, Steiner, Keysers, Uszkoreit et~al.}]{tolstikhin2021mlp}
Ilya~O Tolstikhin, Neil Houlsby, Alexander Kolesnikov, Lucas Beyer, Xiaohua Zhai, Thomas Unterthiner, Jessica Yung, Andreas Steiner, Daniel Keysers, Jakob Uszkoreit, and 1 others. 2021.
\newblock Mlp-mixer: An all-mlp architecture for vision.
\newblock \emph{Advances in neural information processing systems}, 34:24261--24272.

\bibitem[{Van~Kempen et~al.(2024)Van~Kempen, Kim, Tumescheit, Mirdita, Lee, Gilchrist, S{\"o}ding, and Steinegger}]{van2024fast}
Michel Van~Kempen, Stephanie~S Kim, Charlotte Tumescheit, Milot Mirdita, Jeongjae Lee, Cameron~LM Gilchrist, Johannes S{\"o}ding, and Martin Steinegger. 2024.
\newblock Fast and accurate protein structure search with foldseek.
\newblock \emph{Nature biotechnology}, 42(2):243--246.

\bibitem[{Vaswani et~al.(2017)Vaswani, Shazeer, Parmar, Uszkoreit, Jones, Gomez, Kaiser, and Polosukhin}]{vaswani2017attention}
Ashish Vaswani, Noam Shazeer, Niki Parmar, Jakob Uszkoreit, Llion Jones, Aidan~N Gomez, {\L}ukasz Kaiser, and Illia Polosukhin. 2017.
\newblock Attention is all you need.
\newblock \emph{Advances in neural information processing systems}, 30.

\bibitem[{Wang et~al.(2025)Wang, Wang, Jiang, Chen, Shi, and Li}]{wang2025large}
Zhenyu Wang, Zikang Wang, Jiyue Jiang, Pengan Chen, Xiangyu Shi, and Yu~Li. 2025.
\newblock Large language models in bioinformatics: A survey.
\newblock \emph{arXiv preprint arXiv:2503.04490}.

\bibitem[{Xia et~al.(2025)Xia, Jin, Xie, He, Cao, Luo, Liu, Wang, Liu, Chen et~al.}]{xia2025naturelm}
Yingce Xia, Peiran Jin, Shufang Xie, Liang He, Chuan Cao, Renqian Luo, Guoqing Liu, Yue Wang, Zequn Liu, Yuan-Jyue Chen, and 1 others. 2025.
\newblock Naturelm: Deciphering the language of nature for scientific discovery.
\newblock \emph{arXiv preprint arXiv:2502.07527}.

\bibitem[{Xiao et~al.(2024)Xiao, Sun, Jin, Wang, and Wang}]{xiao2024proteingpt}
Yijia Xiao, Edward Sun, Yiqiao Jin, Qifan Wang, and Wei Wang. 2024.
\newblock Proteingpt: Multimodal llm for protein property prediction and structure understanding.
\newblock \emph{arXiv preprint arXiv:2408.11363}.

\bibitem[{Yuan et~al.(2023)Yuan, Shen, Fu, Guan, Ma, Qiao, and Wang}]{yuan2023proteinmae}
Mingzhi Yuan, Ao~Shen, Kexue Fu, Jiaming Guan, Yingfan Ma, Qin Qiao, and Manning Wang. 2023.
\newblock Proteinmae: masked autoencoder for protein surface self-supervised learning.
\newblock \emph{Bioinformatics}, 39(12):btad724.

\bibitem[{Yue and Dill(1992)}]{yue1992inverse}
Kaizhi Yue and Ken~A Dill. 1992.
\newblock Inverse protein folding problem: designing polymer sequences.
\newblock \emph{Proceedings of the National Academy of Sciences}, 89(9):4163--4167.

\bibitem[{Zhang et~al.(2025)Zhang, Li, Luo, Hu, Zhao, Li, Liu, Wang, Bi, Gao et~al.}]{zhang2025unigenx}
Gongbo Zhang, Yanting Li, Renqian Luo, Pipi Hu, Zeru Zhao, Lingbo Li, Guoqing Liu, Zun Wang, Ran Bi, Kaiyuan Gao, and 1 others. 2025.
\newblock Unigenx: Unified generation of sequence and structure with autoregressive diffusion.
\newblock \emph{arXiv preprint arXiv:2503.06687}.

\bibitem[{Zhang and Skolnick(2004)}]{zhang2004scoring}
Yang Zhang and Jeffrey Skolnick. 2004.
\newblock Scoring function for automated assessment of protein structure template quality.
\newblock \emph{Proteins: Structure, Function, and Bioinformatics}, 57(4):702--710.

\bibitem[{Zhang(2016)}]{zhang2016introduction}
Zhongheng Zhang. 2016.
\newblock Introduction to machine learning: k-nearest neighbors.
\newblock \emph{Annals of translational medicine}, 4(11).

\bibitem[{Zhou et~al.(2023)Zhou, Chen, Ye, Wang, Zhang, Mao, Li, Hao, Huang, Tang et~al.}]{zhou2023prorefiner}
Xinyi Zhou, Guangyong Chen, Junjie Ye, Ercheng Wang, Jun Zhang, Cong Mao, Zhanwei Li, Jianye Hao, Xingxu Huang, Jin Tang, and 1 others. 2023.
\newblock Prorefiner: an entropy-based refining strategy for inverse protein folding with global graph attention.
\newblock \emph{Nature Communications}, 14(1):7434.

\end{thebibliography}
\appendix

\section{Appendix}

\subsection{Vocabulary}

The vocabulary used for sequence modeling consists of 23 tokens, as summarized in Table~\ref{tab:vocabulary}. 
This includes the 20 standard amino acids, one unknown token, and two special tokens representing the beginning and end of a protein sequence.

\begin{table}[h!]
\centering
\begin{tabular}{ll}
\toprule
Token & Description \\
\midrule
A & Alanine \\
R & Arginine \\
N & Asparagine \\
D & Aspartic acid \\
C & Cysteine \\
Q & Glutamine \\
E & Glutamic acid \\
G & Glycine \\
H & Histidine \\
I & Isoleucine \\
L & Leucine \\
K & Lysine \\
M & Methionine \\
F & Phenylalanine \\
P & Proline \\
S & Serine \\
T & Threonine \\
W & Tryptophan \\
Y & Tyrosine \\
V & Valine \\
\midrule
X & Unknown token \\
1 & Begin of protein sequence (BOS) token \\
2 & End of protein sequence (EOS) token \\
\bottomrule
\end{tabular}
\caption{Vocabulary used in DS-ProGen.}
\label{tab:vocabulary}
\end{table}

\begin{figure*}[t]
    \centering
    \includegraphics[width=1.0\linewidth]{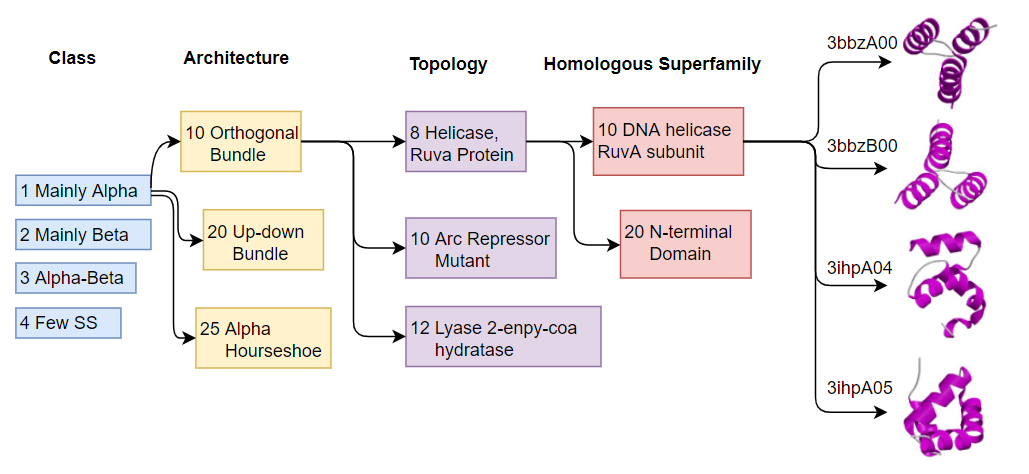} 
    \caption{Topology-based splitting strategy for training set construction.}
    \label{fig:benchmark}
\end{figure*}

\subsection{Benchmark Details} \label{bench}

\subsection*{Training Set Construction}
The training set is constructed from the CATH S35 V4.3 dataset, comprising 32,389 proteins. To enable 10-fold cross-validation, shown in Figure~\ref{fig:benchmark}, we classified proteins based on their topology numbers and split them into 10 disjoint subsets of approximately equal size (3,200--3,400 proteins each). Each split ensures minimal redundancy at both the sequence and topology levels. For each training run, users can randomly select 9 splits for training and the remaining one for validation. 
The protein sequences have lengths ranging from 69 to 15,018 residues, with an average length of 1,163 residues; more than 50\% of the sequences are longer than 1,000 residues. Detailed split lists and scripts for loading specific PDB files are provided in the benchmark package.

\subsection*{Test Set Construction}
The test set is constructed from the CAMEO dataset, selecting 504 protein structures released between July 17, 2021, and March 12, 2022. To ensure independence from the training data, we filter out sequences with high similarity and only retain medium and hard targets. Specifically, we provide three test sets:
\begin{itemize}
    \item \textbf{Full test set:} 504 proteins.
    \item \textbf{Medium+Hard targets:} 336 proteins.
    \item \textbf{Hard-only targets:} 63 proteins.
\end{itemize}
A script is included to auto-update the test set with future CAMEO releases.

\subsection*{Rationale for Dataset Partitioning}
We validate the non-redundancy between the training and test sets by computing sequence similarity scores using the BLOSUM62 substitution matrix and structural similarity scores using TM-score. The average sequence similarity between the training and test datasets is 0.1844, with a maximum below 0.21, confirming substantial dissimilarity between them.

\subsection*{Comparison with Existing Datasets}
Compared to traditional datasets such as CATH, SCOP, and CASP, our split strategy eliminates redundancy and prevents data leakage. Unlike datasets focusing solely on structural quality, our benchmark also emphasizes sequence diversity.

\begin{figure*}[h]
    \centering
    \begin{subfigure}{1.0\textwidth}
        \centering
        \includegraphics[width=\linewidth]{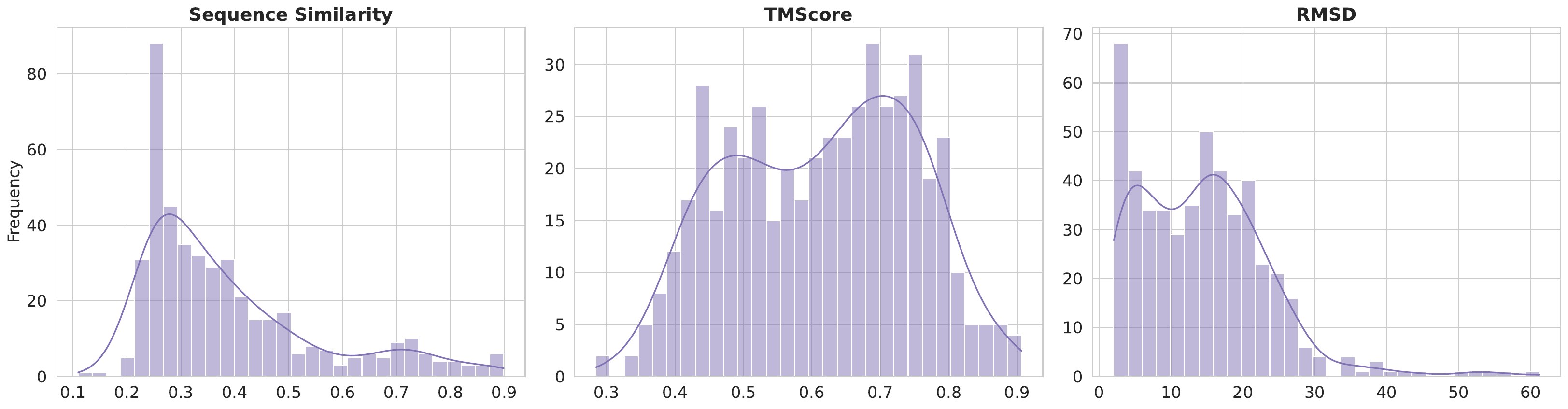}
        \caption{Distribution between the test set and the pretraining set(4M).}
        \label{fig:data_analysis_4m}
    \end{subfigure}
    
    \vspace{1em}

    \begin{subfigure}{1.0\textwidth}
        \centering
        \includegraphics[width=\linewidth]{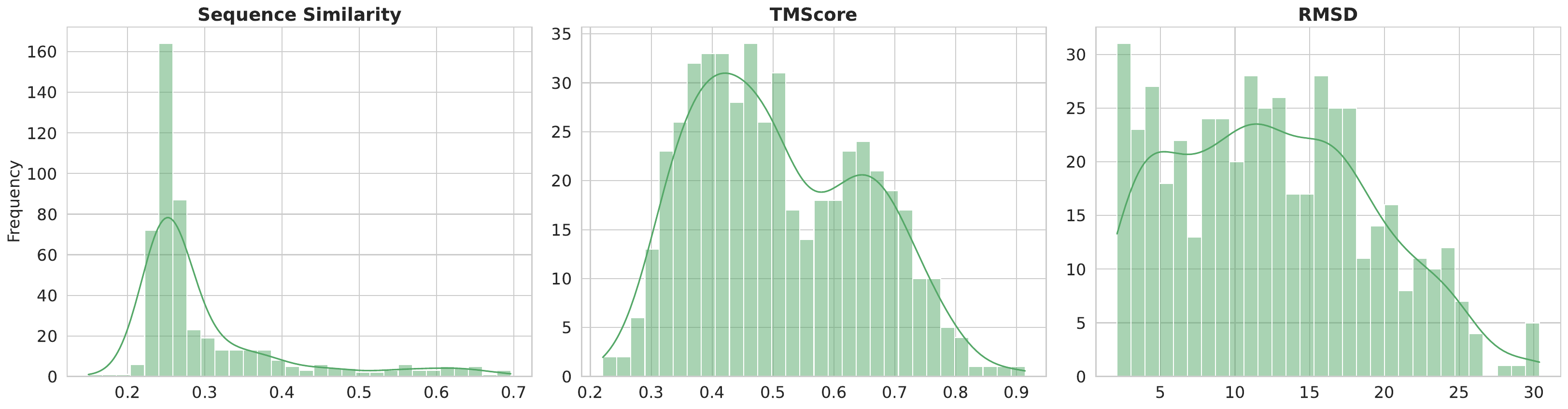}
        \caption{Distribution between the test set and fine-tuning set (30k).}
        \label{fig:data_analysis_29k}
    \end{subfigure}
    
    \vspace{1em}

    \begin{subfigure}{1.0\textwidth}
        \centering
        \includegraphics[width=\linewidth]{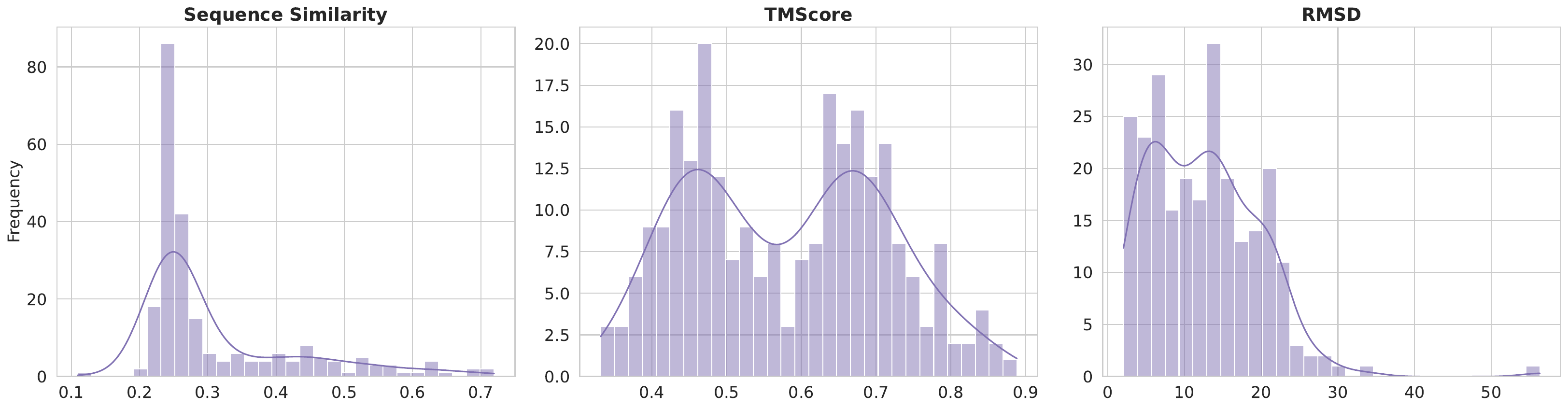}
        \caption{Distribution between the test set and the pretraining set with surface features (80k).}
        \label{fig:data_analysis_80k}
    \end{subfigure}
    
    \caption{
Distribution of maximum sequence similarity, maximum TM-Score, and minimum RMSD between each test protein and the nearest neighbor in different training datasets. The results confirm that the test set is sufficiently distinct from the training datasets, with no evidence of data leakage.
}
\label{fig:data_analysis}
\end{figure*}

\subsection{Data Analysis}

To ensure fair evaluation and eliminate potential data leakage, we conducted a comprehensive data analysis comparing the test set against three different training datasets:
(1) all of the pretraining dataset,
(2) all of the benchmark training dataset,
(3) pretraining dataset with surface features.

For each test protein, we identified the most similar training sample based on sequence or structural similarity: specifically, we recorded the maximum sequence similarity (via BLAST \cite{camacho2009blast+}) and TM-Score (via Foldseek \cite{van2024fast}) as well as the minimum RMSD.

It is worth noting that due to the approximate nature of many-to-many comparison (rather than exhaustive all-to-all search), slight matching inaccuracies may exist. However, overall trends remain reliable for evaluating potential overlaps.

The analysis results demonstrate that across all datasets, maximum sequence similarities are generally low (mostly below 0.5), and most of the maximum TM-Scores are under 0.6, indicating limited sequential and structural similarity between the test set and training data. Minimum RMSD values are relatively high, often exceeding 5 Å, further supporting the separation. No significant data leakage is observed, ensuring the validity of the evaluation.

Further comparison between different datasets reveals important trends. The 4M and 80K pretraining datasets show broad diversity while maintaining low similarity to the test set, providing a strong foundation for the model to learn generalized sequence-structure relationships and validating the necessity of pretraining. These observations collectively confirm that both pretraining and fine-tuning are essential to achieving robust model performance.

\subsection{Results}

\begin{figure}[H]
    \centering
    \includegraphics[width=\linewidth]{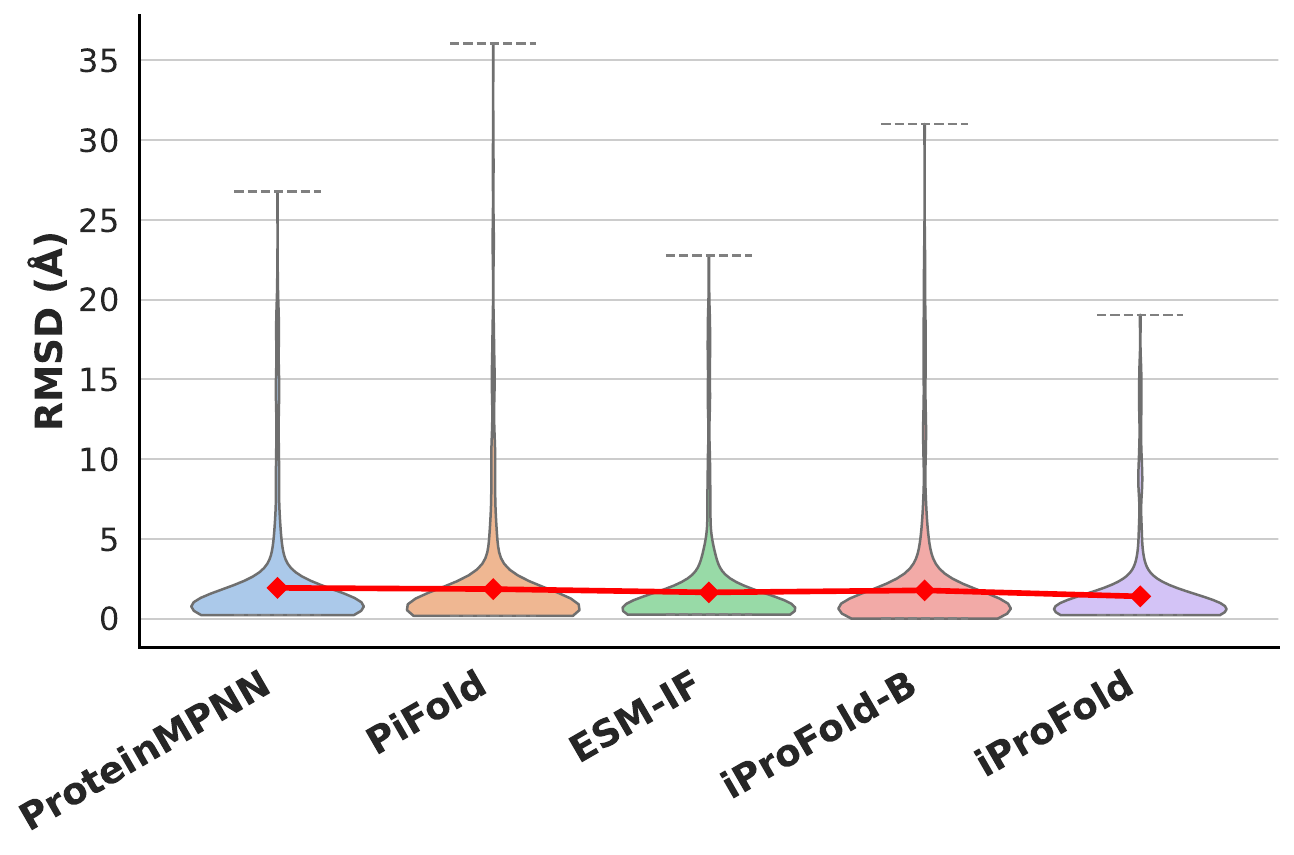}
    \caption{Distribution of RMSD (\AA) on PRIDE test set across baselines and our models, including statistical markers for both the mean and median RMSD (\AA).}
    \label{fig:recovery_rate_violin}
\end{figure}

\begin{figure}[H]
    \centering
    \includegraphics[width=\linewidth]{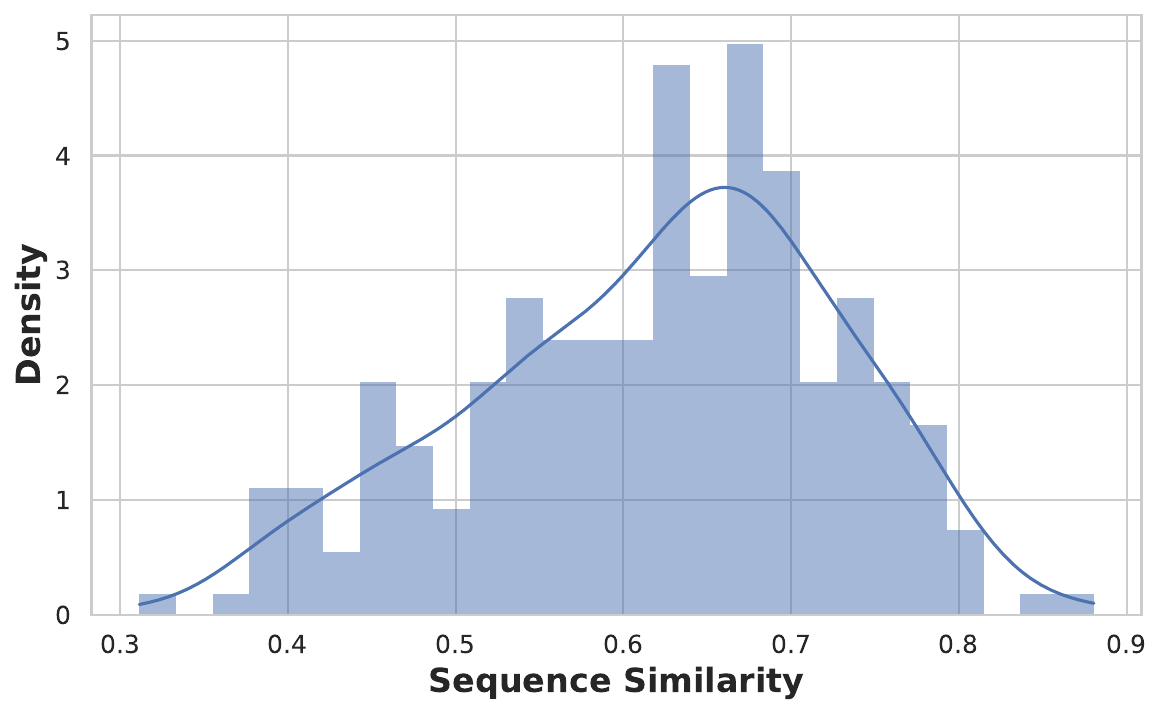}
    \caption{Histogram and kernel density estimation (KDE) plot of sequence similarity scores between generated sequence and groundtruth on PRIDE test set.}
    \label{fig:seq_sim}
\end{figure}

\begin{figure*}[t]
    \centering
    \begin{subfigure}{0.48\textwidth}
        \centering
        \includegraphics[width=\linewidth]{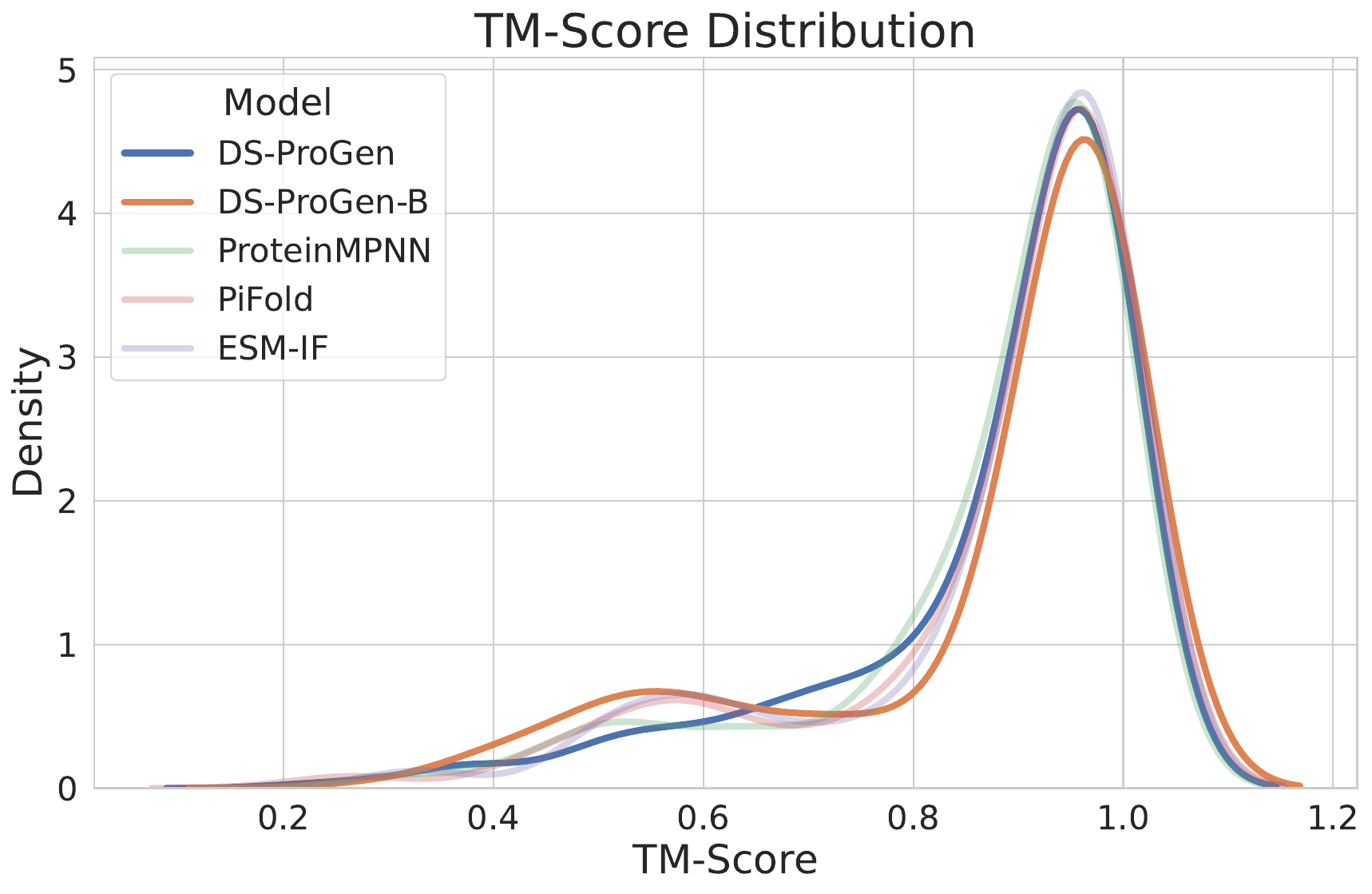}
        \caption{}
        \label{fig:tmscore_violin}
    \end{subfigure}
    \hfill
    \begin{subfigure}{0.48\textwidth}
        \centering
        \includegraphics[width=\linewidth]{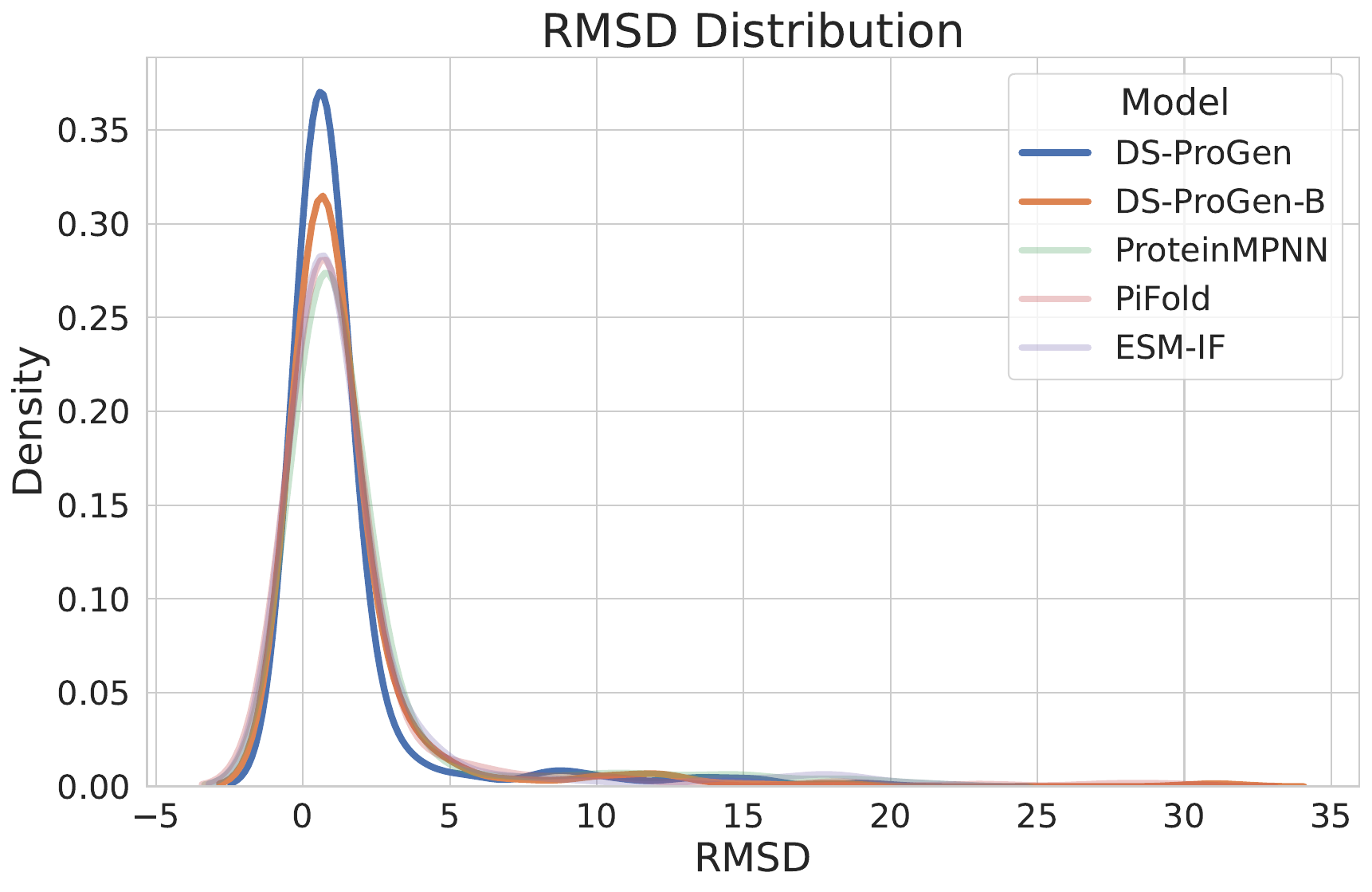}
        \caption{}
        \label{fig:rmsd_violin}
    \end{subfigure}
    \caption{Kernel density estimation plots for structures predicted by AlphaFold3 from model-generated sequences, comparing the structural quality of designed protein sequences across baselines and our models. (a) shows the distribution of TMScore values and (b) illustrates RMSD distributions.}
    \label{fig:tmscore_rmsd_violin}
\end{figure*}

\subsection{Additional Case Studies}

\begin{figure*}[t]
    \centering
    \includegraphics[width=\textwidth]{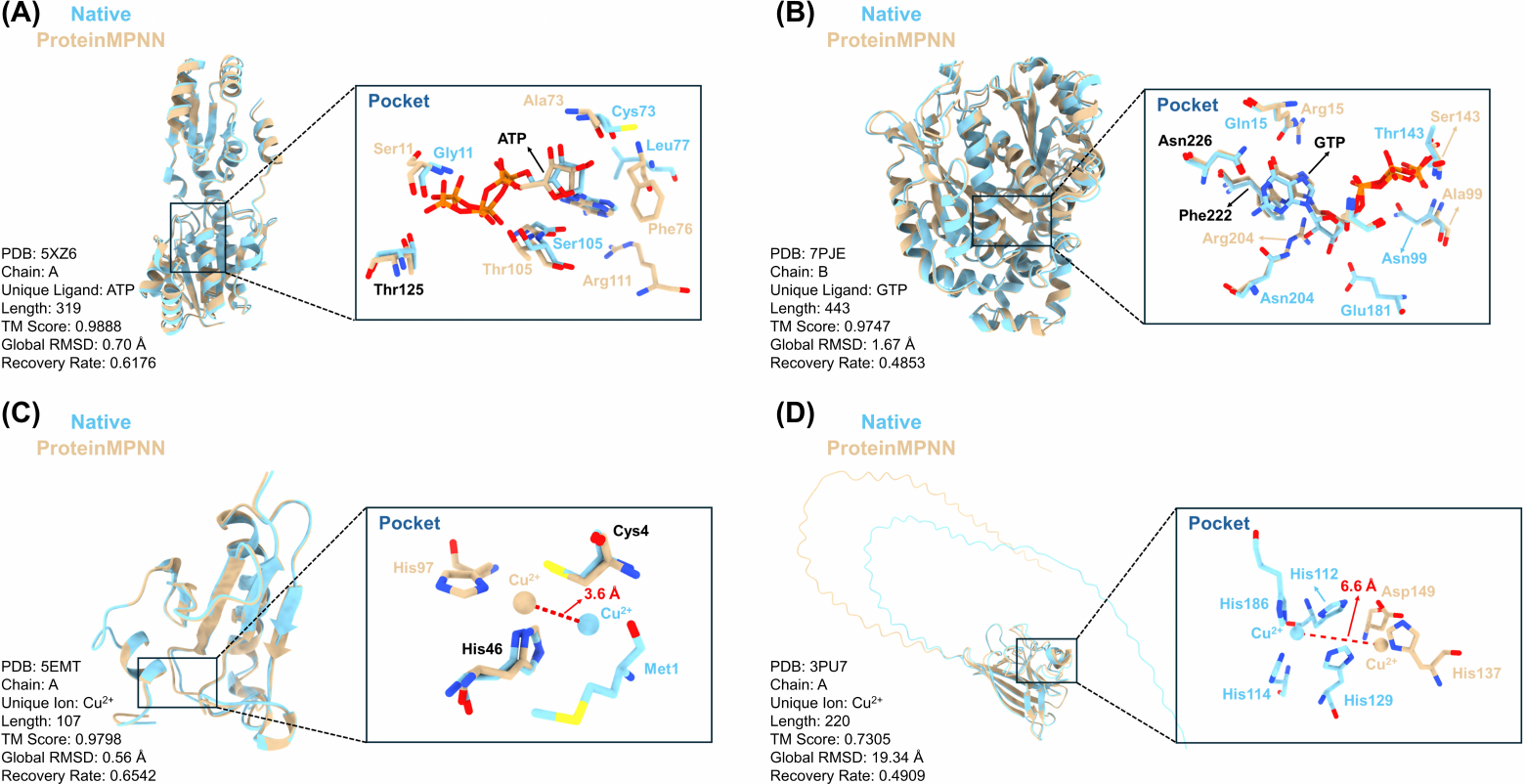}
    \caption{Structural alignment between ground truth structures (blue) and predicted structures (light brown) folded from protein sequences generated by ProteinMPNN, for the same representative cases used in the DS-ProGen visualization. Each zoomed-in panel highlights the corresponding binding pocket suited for a specific ligand or ion.}
    \label{ProteinMPNN}
\end{figure*}

\begin{figure*}[h]
    \centering
    \includegraphics[width=\textwidth]{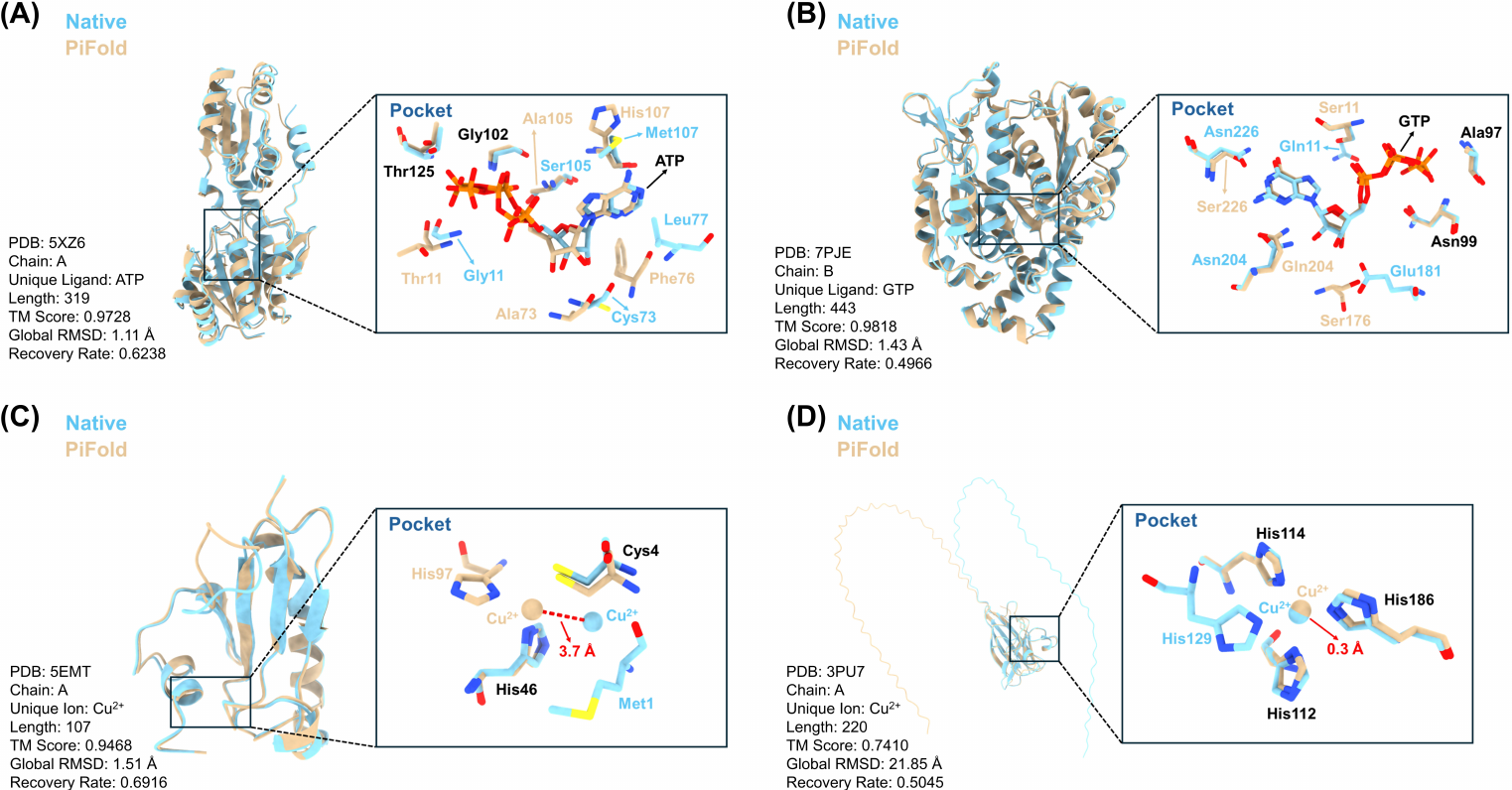}
    \caption{Structural alignment between ground truth structures (blue) and predicted structures (light brown) folded from protein sequences generated by PiFold, for the same representative cases used in the DS-ProGen visualization. Each zoomed-in panel highlights the corresponding binding pocket suited for a specific ligand or ion.}
    \label{PiFold}
\end{figure*}

\begin{figure*}[t]
    \centering
    \includegraphics[width=\textwidth]{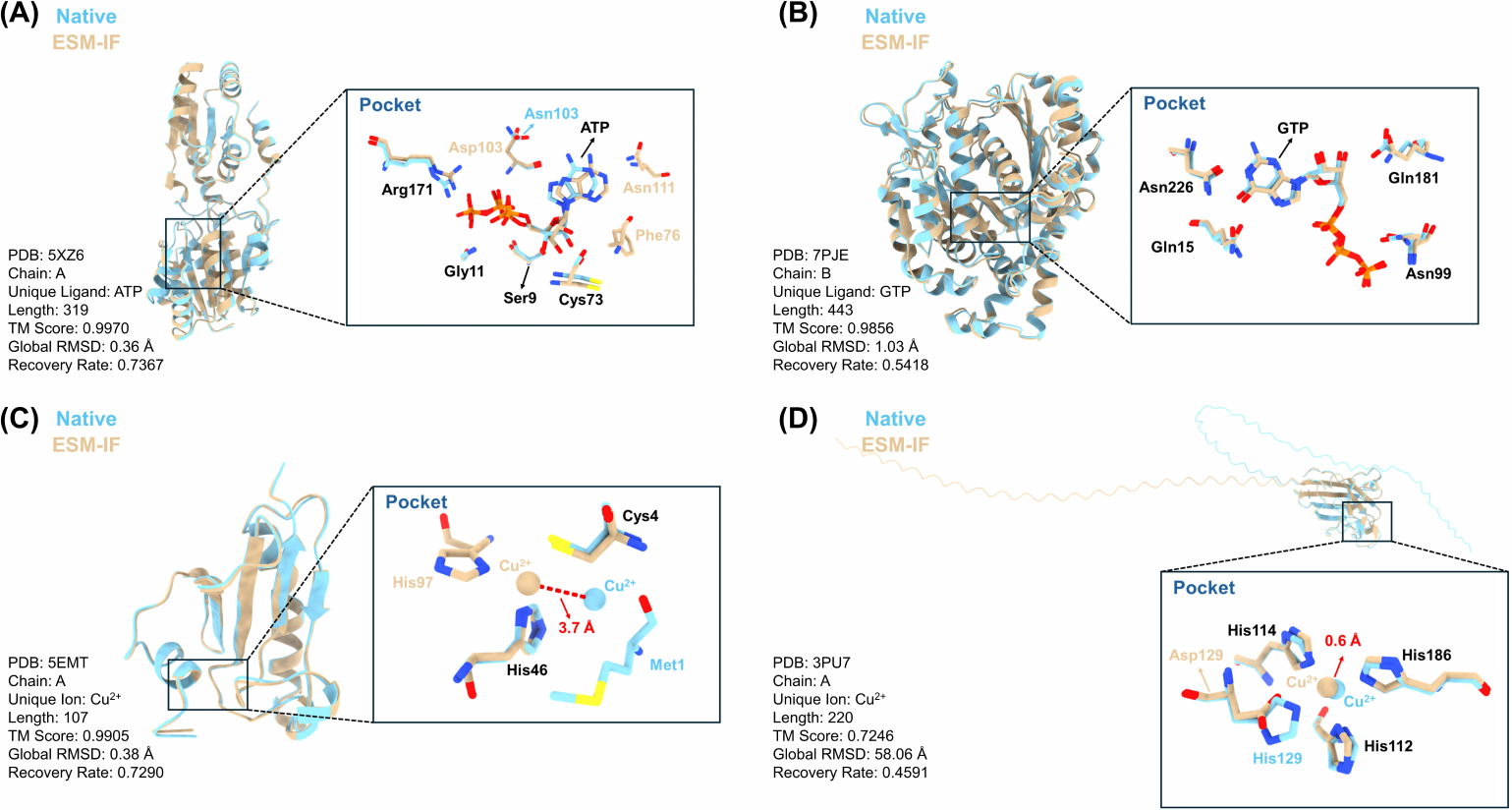}
    \caption{Structural alignment between ground truth structures (blue) and predicted structures (light brown) folded from protein sequences generated by ESM-IF, for the same representative cases used in the DS-ProGen visualization. Each zoomed-in panel highlights the corresponding binding pocket suited for a specific ligand or ion.}
    \label{ESM-IF}
\end{figure*}

\end{document}